\documentclass{WileyMSP-template}

\usepackage{amsmath,amsfonts}
\usepackage{algorithmic}
\usepackage{algorithm}
\usepackage{array}
\usepackage[caption=false,font=normalsize,labelfont=sf,textfont=sf]{subfig}
\usepackage{textcomp}
\usepackage{stfloats}
\usepackage{url}
\usepackage{verbatim}
\usepackage{graphicx}
\usepackage{cite}
\usepackage{xcolor}
\usepackage{hyperref}
\usepackage{xurl}
\usepackage{placeins}
\newif\ifshowquestions
\newif\ifshowrevision
\showrevisionfalse

\newcommand{\revision}[1]{%
  \ifshowrevision\textcolor{red}{#1}\else#1\fi}

\newcommand{\question}[1]{%
  \ifshowquestions\textcolor{blue}{\textbf{Q: #1}}\fi}

\newcommand{\MaxAccNoV}{$95.89\%$}
\newcommand{\MaxAccVFive}{$95.3\%$}
\newcommand{\MaxAccVTwenty}{$94.2\%$}
\newcommand{\MaxAccNoVOneBit}{92.52\%}
\newcommand{\MaxAccNoVTwoBit}{$94.8\%$}

\begin{document}

\pagestyle{fancy}
\rhead{\includegraphics[width=2.5cm]{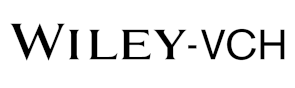}}

\title{On the Role of Preprocessing and Memristor Dynamics in Reservoir Computing for Image Classification}
\maketitle

\author{Rishona Daniels\textsuperscript*}
\author{Duna Wattad\textsuperscript}
\author{Ronny Ronen\textsuperscript}
\author{David Saad\textsuperscript}
\author{Shahar Kvatinsky\textsuperscript*}

\begin{affiliations}
Rishona Daniels, Duna Wattad, Ronny Ronen, and Prof. Shahar Kvatinsky \\
Viterbi Faculty of Electrical and Computing Engineering, Technion - Israel Institute of Technology, Haifa, Israel\\
Email Address: rishonad@campus.technion.ac.il, shahar@ee.technion.ac.il

Prof. David Saad \\
College of Engineering and Physical Sciences, Aston University, Birmingham B4 7ET, United Kingdom
\end{affiliations}


% Abstract should be written in the present tense and impersonal style (i.e., avoid we), and be at most 200 words long

\begin{abstract}
Reservoir computing (RC) is an emerging recurrent neural network architecture that has attracted growing attention for its low training cost and modest hardware requirements. Memristor-based circuits are particularly promising for RC, as their intrinsic dynamics can reduce network size and parameter overhead in tasks such as time-series prediction and image recognition. Although RC has been demonstrated with several memristive devices, a comprehensive evaluation of device-level requirements remains limited. 
In this paper, we analyze and explain the operation of a parallel delayed feedback network (PDFN) RC architecture with volatile memristors, focusing on how device characteristics -- such as decay rate, quantization, and variability -- affect reservoir performance. We further discuss strategies to improve data representation in the reservoir using preprocessing methods and suggest potential improvements.
The proposed approach achieves \MaxAccNoV\ classification accuracy on MNIST, comparable with the best reported memristor-based RC implementations. Furthermore, the method maintains high robustness under $20\%$ device variability, achieving an accuracy of up to \MaxAccVTwenty. 
These results demonstrate that volatile memristors can support reliable spatio-temporal information processing and reinforce their potential as key building blocks for compact, high-speed, and energy-efficient neuromorphic computing systems.
\end{abstract}

\section{Introduction}
\label{sec:intro}
\question{
Para 1: Why is RC needed?}

Deep learning has driven remarkable progress across domains such as natural language processing, computer vision, and control systems. However, these models require substantial computational resources and training time, posing significant challenges in terms of energy consumption and hardware requirements. Reservoir computing (RC)~\cite{Jeager_2001_echo_state_approach, Maass_2003_LSM} has emerged as a promising alternative for energy-efficient data processing. In RC, a fixed dynamical system called the~\emph{reservoir} projects inputs into a high-dimensional space, and only a single \emph{readout} layer is trained, typically with simple linear regression. This separation enables efficient training and reduces hardware complexity compared to conventional recurrent neural networks.

\question{Para 2: What can theoretical RC do?}

RC has been applied for a wide range of tasks, including speech and audio recognition \cite{photonic_rc_for_speech_recognition, hopf_physical_rc_for_sound_recognition}, seizure and epileptic activity detection from EEG and iEEG data \cite{rc_for_epileptic_seizures}, robotics \cite{physical_rc_for_robotics}, wind speed prediction \cite{wind_forecasting_using_rc}, stock market prediction \cite{stock_market_prediction_using_rc}, and image and video recognition \cite{img_recog_using_rc, photonic_rc_for_human_action_recognition}. 
RC has also been used in time-series prediction tasks, including applications involving chaotic dynamics~\cite{ngrc_for_chaotic_systems_prediction}.
%Particularly, RC has proven highly effective for time-series prediction, especially in systems with chaotic dynamics~\cite{ngrc_for_chaotic_systems_prediction}.   

\question{Para 3: Why and how to do physical RC (High level)?}

A key strength of RC is its compatibility with physical substrates. Since only the readout layer requires training, the reservoir can be realized by exploiting the intrinsic dynamic properties of various physical systems~\cite{experimental_unification_of_rc}. Examples include photonic systems, which provide ultrafast optical interference and delay dynamics;\cite{photonic_rc}, electronic systems, which leverage device and circuit nonlinearities and charge-transport effects in mixed-signal or analog implementations\cite{physical_rc_with_emerging_electronics}; mechanical systems, where deformations and vibrations generate rich temporal responses used for control\cite{physical_rc_with_origami}, and spintronic systems which use fast changes in magnetization controlled by electric signals, enabling compact and low-power computation\cite{spintronic_rc}. These media naturally provide the nonlinear, high-dimensional, and fading-memory dynamics essential for RC. Leveraging such dynamics enables ultrafast computation at the native timescale of the medium (e.g., speed of light for photonics, GHz signals in electronics), while consuming far less energy than digital implementations. Physical reservoirs can also potentially offer an inherent massive parallelism, making RC an attractive platform for unconventional, energy-efficient, and high-speed computing.

\question{Para 4: Why is RC with memristors good?}

Among these various physical media, memristive reservoir computing has emerged as an especially promising approach for hardware-efficient neural computation~\cite{in_memory_in_sensor_rc_with_memristors_review}. Memristors, and particularly volatile memristors, intrinsically exhibit the two key properties required for reservoir computing: nonlinear dynamics and short-term memory. When write pulses are applied, the internal state of a volatile memristor changes in a nonlinear manner and subsequently relaxes toward its equilibrium value due to the device’s inherent volatility. These properties enable a simplified implementation of reservoirs that can project input signals into a high-dimensional space directly within the hardware~\cite{rc_with_dynamic_memristors}. Here, high-dimensional state space refers to the large number of distinguishable internal responses produced when nonlinear state updates interact with fading memory, causing even small differences in input sequences to generate diverging transient states~\cite{lukosevicius_jaeger_2009,jaeger_haas_2004}. The memristors' nanoscale dimensions, integration with CMOS technology, and compatibility with crossbar architectures enable dense, scalable reservoirs, while their low power consumption and nanosecond-scale operation support real-time processing of complex temporal signals \cite{rc_with_self_organizing_nanowire_memristors}. These advantages make memristive RC well-suited for energy-efficient, high-speed applications in time-series prediction, signal processing, and image recognition.

\question{Para 6: What is lacking in the literature, and what do we do in this paper? What do we get?
}

Several prior works have demonstrated memristive RC using a variety of different methods; however, the effect of practical considerations like device quantization and variability remains insufficiently characterized. 

In this paper, we present a comprehensive analysis of reservoir computing (RC) implemented with volatile memristors, focusing on how device-level dynamics shape system-level computational performance. Building upon the inherent nonlinear and fading-memory behavior of volatile devices, we employ a popular (yet unnamed) method in literature, we term parallel delayed feedback network (PDFN) architecture \cite{rc_with_dynamic_memristors} that enables multiple memristors to operate in parallel, emulating temporal dependencies without crossbar interconnections. This configuration combines the temporal richness of delayed feedback networks with the scalability of parallel reservoirs, providing an efficient and hardware-feasible platform for physical RC.

The main contributions of this paper are:
\begin{enumerate}
    \item Deeper insights and evaluation of PDFN architecture for the image recognition task. 
    \item Comprehensive evaluation of the impact of preprocessing methods, including sectioning and parity transformations, that enhance image-based RC performance.
    \item Systematic analysis of how decay rate, quantization, and device variability affect RC performance in a spatio-temporal (MNIST) task.
    \item  Demonstration of high performance, achieving \MaxAccNoV\ accuracy on MNIST, comparable to the best reported accuracy of prior memristive RC works, with robustness maintained even under $20\%$ device variability.
\end{enumerate}

\question{Para 7: Structure of paper}

The rest of the paper is organized as follows. Section~\ref{sec:background} provides background on memristive RC,
Section~\ref{sec:image_recognition} explains the image recognition task using PDFN, including system design and preprocessing.
Section~\ref{sec:evaluation_methodology} gives a detailed evaluation methodology of the experiments performed to understand the impact of memristor parameters, and Section~\ref{sec:results} discusses the evaluation results. 
Finally, Section~\ref{sec:conclusion} concludes the paper. 

\section{Background}
\label{sec:background}

The following section explains the theoretical framework of reservoir computing and its types, with emphasis on the differences between the respective methods. Next, memristors are introduced, focusing specifically on their properties that can be beneficial for physical reservoir computing. Then, we explain how the properties of memristors can be used to build physical RC systems. 

\subsection{Reservoir Computing}
\label{subsec:reservoir_computing}

\question{Para 1: What is RC?}

\begin{figure}[H]
\centering 
\includegraphics[scale=0.33]{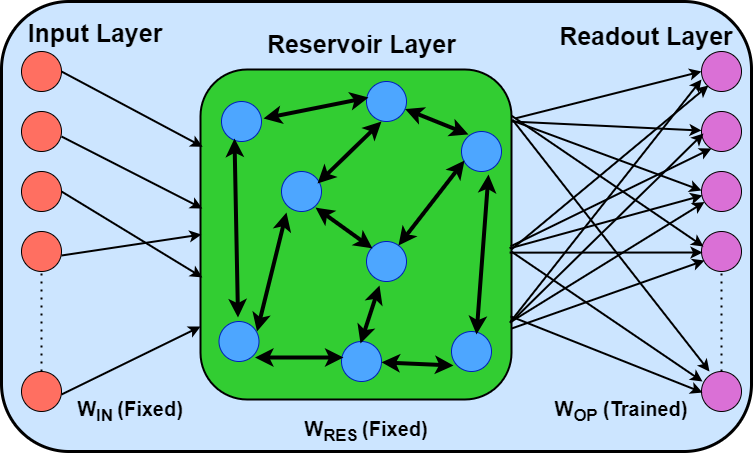}
\caption{Block diagram of RC with input, reservoir, and readout layers. Only readout weights ($W_{OP}$) are trained; input ($W_{IN}$) and reservoir parameters ($W_{RES}$) are not trained.}
\label{fig_RC_Blk_diag}
\end{figure}

Reservoir computing is a computational framework that uses the transient dynamics of a system to process information.
It is a type of recurrent neural network comprising three parts: an input layer, a reservoir layer, and a readout layer (see Figure \ref{fig_RC_Blk_diag})~\cite{Jeager_2001_echo_state_approach}. The input layer encodes the input data into a time-varying signal and applies it to the reservoir. The reservoir is a dynamical system (a system that evolves with time) that projects the input sequence into a high-dimensional state space. The reservoir must have two related properties: the echo-state property and the fading memory property. The \textit{echo state property} ensures that the reservoir’s current state depends mainly on recent inputs rather than its initial conditions, while the \textit{fading memory property} means that the influence of past inputs gradually diminishes over time. The trainable readout layer then interprets this representation. Broadly, RC can be categorized into three types: echo state networks (ESN)~\cite{Jeager_2001_echo_state_approach}, liquid state machines (LSM)~\cite{Maass_2003_LSM}, and delayed feedback networks (DFN)~\cite{info_processing_in_single_dynamical_system}. In all three RC types, training is confined to the readout, which keeps learning simple and hardware-friendly.

\question{Para 2: What is ESN?}

\begin{figure}[H]
\centering 
\includegraphics[scale=0.65]{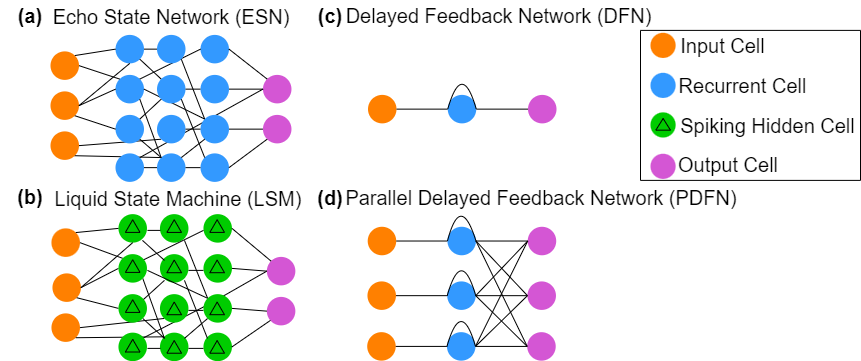}
\caption{Network structures of (a) echo state network (ESN), (b) liquid state machine (LSM), (c) delayed feedback network (DFN), and (d) parallel delayed feedback network (PDFN). Figure adapted from \cite{neural_network_zoo}}
\label{all_RC_types_block_diagram}
\end{figure}

An echo state network (ESN) implements the reservoir as a large, sparsely and randomly connected recurrent neural network of simple nonlinear units \cite{Jeager_2001_echo_state_approach} (see Figure \ref{all_RC_types_block_diagram}(a)). 
%These units are often leaky integrators to match input time scales. 
Each unit acts as a simple dynamic element and can often be modeled as a leaky integrator. 
Leaky integrators, by definition, possess the echo-state and the fading memory properties needed for RC. 
ESNs operate in discrete time and train only the readout, making them efficient for tasks such as time-series prediction.

\question{Para 3: What is LSM?}

Liquid State Machines (LSM) realize the reservoir using spiking neurons, capturing temporal information through biologically inspired dynamics \cite{Maass_2003_LSM} (see Figure \ref{all_RC_types_block_diagram}(b)). The “liquid” metaphor emphasizes how input perturbations ripple through the network, producing rich spatio-temporal spike patterns in continuous time. As in ESNs, only the readout layer is trained, but LSMs explicitly leverage event-driven dynamics, which is advantageous for neuroscience-inspired computation and for processing event-based sensor data.

\question{Para 4: What is DFN?}

Delayed Feedback Networks (DFN) replace a large recurrent reservoir with a single nonlinear node and a delay line feedback \cite{info_processing_in_single_dynamical_system}. 
Time-multiplexing along the delay line creates a sequence of states of the node in time, yielding a high-dimensional state without a large physical network (see Figure \ref{all_RC_types_block_diagram}(c)). 
This differs from ESN, in which the input data is spatially distributed across several reservoir nodes. The DFN architecture is attractive in hardware implementations, such as photonic or electronic reservoirs, because it provides rich dynamics with minimal circuit complexity. In this paper, we focus on an extended version of DFN which comprises several DFNs working simultaneously with a common readout (see Figure \ref{all_RC_types_block_diagram}(d)). We call this architecture \textit{parallel delayed feedback network (PDFN)}, which is described in Section \ref{subsec:rc_with_memristors}. 

\question{Para 5: Common parameters that constitute a good reservoir }

A good reservoir is defined by a set of parameters that determine how effectively it transforms the input into a rich, interpretable state space for the readout layer~\cite{experimental_unification_of_rc}. 
The reservoir size (the number of nodes) controls the dimensionality of the state space and thus the richness of the representation. 
The memory characteristics, governed by the leak or decay rate of the leaky integrator, determine how long past inputs influence the current state.
The nonlinearity of the nodes allows the system to transform inputs in a way that enhances class separation or feature extraction.
The input scaling regulates the strength of the input signal entering the reservoir, affecting its dynamic range and stability. 
Different applications depend on these properties in different ways. For example, tasks that involve predicting future values from past data often need the system to retain information for a longer time. Tasks that classify single, static inputs instead depend more on the system's ability to separate patterns in a nonlinear way. Meanwhile, tasks that must recognize sequences or events unfolding over time require a combination of both memory and nonlinear processing.
Because capacity is finite, there is a tradeoff - pushing memory too long can reduce effective nonlinearity (and vice versa). 
  
\question{Para 6: Why do RC in neuromorphic hardware?}

As described above, reservoir computing requires complex reservoir dynamics.  
The echo-state and fading memory properties are resource-intensive and require extensive memory updates and data movement in traditional digital computers, and are more suitable for implementations utilizing physical properties of hardware. Reservoir computing implemented on neuromorphic hardware can enable in-memory information processing directly in the physical domain in a straightforward and elegant manner, as we show further in this paper.

Among the various neuromorphic hardware, memristors are particularly interesting, as they natively possess the key properties needed for reservoir computing. This enables physical RC with fewer components and enables energy-efficient computing. The properties of memristors are described in the next section. 

\subsection{Memristors}
\label{subsec:memristors}

%\question{Para 1: Intro para to chua's memristor}

\question{Para 1: What is a memristor? How to classify? Equations?}

For neuromorphic RC, we utilize memristors in this paper. A memristor is a two-terminal device whose conductance changes with electrical stimulus and can encode information as a resistive level. The stored state can be retrieved by applying a small, non-destructive voltage or current, enabling reliable readout without altering the memory content. 
Since the concept of memristor was first introduced by Leon Chua in 1971 \cite{Chua_1971} and the first nanoscale experimental demonstrations in 2008 \cite{Strukov_2008}, various memristive device families with distinct properties have emerged.

A memristive device \cite{Chua_Kang_1976, Chua_2011, Chua_2014} is defined by:

\begin{equation}
    \frac{dx}{dt} = f(x, u, t),
\end{equation}

\begin{equation}
y = g(x, u, t) \cdot u,
\end{equation}
where $x$ is the internal state (dimensionless), $t$ is time, $u$ is the input (voltage or current), and $y$ is the output (current or voltage). 

\question{Para 2: Volatile vs nonvolatile memristors}

If $f(x, 0, t) = 0$ for any $x$ and $t$, the memristor is \textit{nonvolatile} and holds the resistance/state without bias.
If $f(x, 0, t) \neq 0$ for some $x$ and $t$, the state drifts in time even without input, i.e., the device is \textit{volatile}.
Several volatile memristors have been fabricated and experimentally characterized in the literature. The volatility of these devices ranges from seconds~\cite{optically_controlled_MoS2_sec} and milliseconds~\cite{flexible_TiO2_WOx3_memristor} to a few hundred picoseconds~\cite{picosecond_VO2_memristors}.
Volatile memristors can be modeled as leaky integrators and, therefore, naturally implement the fading-memory property needed for the reservoir in an RC system. Nonvolatile memristors are well-suited for storing trained readout weights.

\question{Para 3: Types of memristor models in literature and what do we use?}
%Memristors are commonly classified by retention (volatile vs. nonvolatile), state granularity (digital vs. analog), and switching polarity (unipolar vs. bipolar).
Several memristor models have been proposed in the literature; they can be broadly divided into generic models and physical models. 
Generic models approximate device behavior under various input conditions and have tunable parameters to fit various physical devices. Physics-based models capture the specific behavior of a device on the basis of the physical operating mechanism. For example, they can capture filament dynamics, ionic diffusion, or threshold switching mechanisms. Generic models are easier to simulate and more flexible, but less accurate; physical models are more accurate, but computationally heavier to simulate and have limited flexibility. VTEAM \cite{VTEAM} with short-term memory \cite{VTEAM_with_STM} and VVTEAM \cite{VVTEAM} are examples of generic volatile memristor models, while dynamic memristor \cite{model_of_Wox_memristor} and diffusive memristor \cite{diffusive_memristor} are examples of physical volatile memristor models. 
In this paper, we model RC using the dynamic memristor model.  

%\subsubsection{Dynamic Memristor}

\begin{comment}

\begin{figure*}[!t]
\centering 
\includegraphics[scale=0.35]{Concurrent_dynamic_memristor_internal_state.png}
\caption{State evolution of dynamic input for varying input pulse trains}
\label{dynamic_memristor_state_evolution}
\end{figure*}
\end{comment}

\question{Para 4: Intro to dynamic memristors, dx/dt and I equations}

The dynamic memristor is described by the following expressions:

\begin{equation}
    \frac{dx}{dt} = \lambda \cdot sinh(\eta \cdot V) - \frac{x}{\tau},
    \label{dx/dt}
\end{equation}

\begin{equation}
I = (1 - x) \cdot \alpha \cdot [1 - exp(-\beta \cdot V)] + x \cdot \gamma \cdot sinh(\delta \cdot V),
\end{equation}
where $x \in [x_{Min}, x_{Max}]$ is the internal state variable, $V$ and $I$ are the applied voltage and the resulting current, $\alpha$, $\beta$, $\gamma$, $\delta$, $\lambda$, and $\eta$ are positive fitting parameters dependent on material and device properties, and $\tau$ is the diffusion time constant that determines the rate of decay of $x$. The values of the constants are listed in Table \ref{table:dynamic_memristor_parameters}.

\question{Para 5: Simplified delta x equations}

\begin{figure}[H]
  \centering
\includegraphics[scale=0.44]{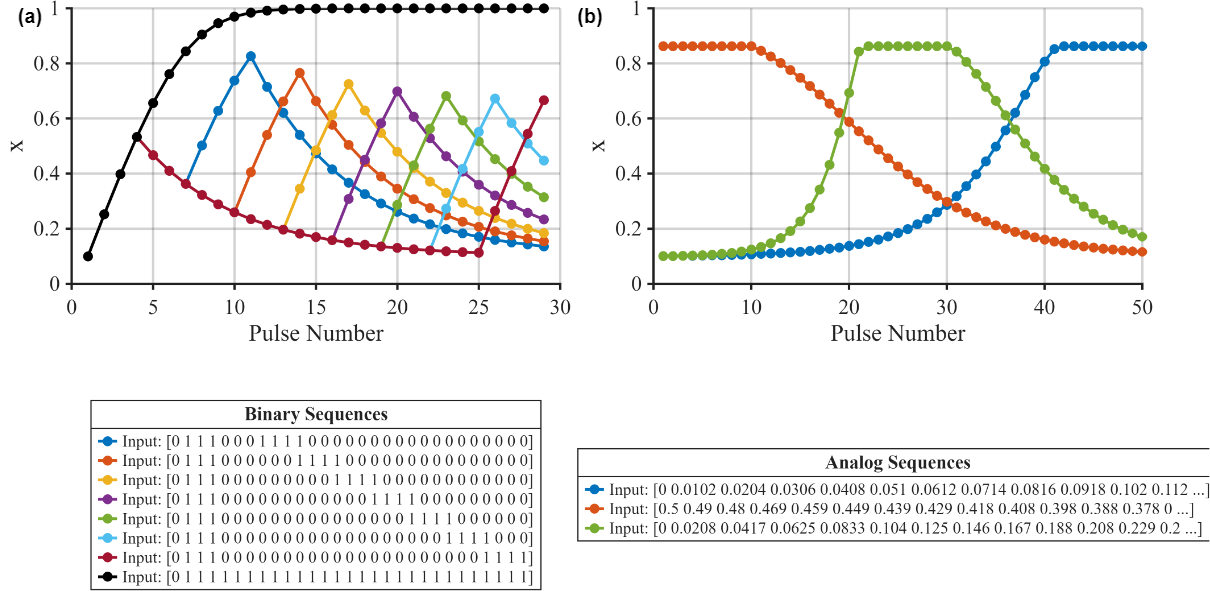}
  \caption{Evolution of internal state variable $x$ of the dynamic memristor (based on (\ref{w_update})-(\ref{w_decay})) under different input pulse trains: (a) binary inputs where the applied voltage alternates between $1.5V$ (logic ‘1’) and $0V$ (logic ‘0’); and (b) analog inputs where the applied voltage varies continuously between $0.8V$ and $1.8V$ (input sequence has values between 0 and 0.5 and are scaled to the write voltage range of the memristor). Each curve represents the device response to a distinct input sequence, illustrating the effect of pulse amplitude modulation on the memristor state dynamics. The model was simulated using MATLAB.}
  \label{memristor_internal_state}
\end{figure}

For analysis and discrete-time simulation, (\ref{dx/dt}) is often approximated during a positive write pulse (of width $t_{pulse}$ and amplitude $V_{pulse}$ $>$ $V_{th}$) as

%When a positive write pulse ($Vpulse> 0.6V$) is applied across the memristor, Equation~(\ref{dx/dt}) can be approximated~\cite{rc_with_dynamic_memristors} to 
\begin{equation}
    \Delta x = R(x)\cdot t_{pulse} \cdot \lambda \cdot sinh(\eta \cdot V_{pulse}),
    \label{w_update}
\end{equation}
\begin{equation}
    R(x) = 1 - \frac{exp(3 \cdot x)}{exp(3 \cdot x_{Max})},
    \label{R(x)}
\end{equation}
where $R(x)$ is the window function that constrains the value of $x$, and $V_{th}$ is the threshold voltage, below which the internal state is not modified. 
Between pulses or during low-bias ($V_{pulse} < V_{th}$), volatility induces leakage toward $x_{\min}$; a first-order discretization yields
%Conversely, if no write pulse is applied ($Vpulse < 0.6V$) across the memristor, Equation~(\ref{dx/dt}) is approximated to reflect the volatile nature of the memristor such that $x$ decays: 
\begin{equation}
        \Delta x = (x - x_{Min}) \cdot \left(1 -  exp\left(-\frac{t_{pulse}}{\tau}\right)\right).
    \label{w_decay}
\end{equation}
i.e., $x$ decays exponentially in the absence of sufficient stimulus. 
If the applied voltage, $V_{pulse}$, is greater than $V_{th}$, the internal state updates according to (\ref{w_update}), and if the applied voltage is lower than $V_{th}$, the internal state decays according to (\ref{w_decay}).
Figure \ref{memristor_internal_state} shows the dynamic evolution of the internal state of the dynamic memristor according to (\ref{w_update}) and (\ref{w_decay}) in response to varying digital and analog stimuli. Different sequences and amplitudes of input signals cause the memristor state to evolve uniquely, leading to a distinct final representation of the input in the memristor~\cite{rc_with_dynamic_memristors}. 
We use the value of $\tau$ as a value ranging from $1 ns$ to $50 ns$. 

Figure \ref{memristor_internal_state}(a) shows the evolution of the internal state variable $x$ for varying binary input conditions. When write pulses with amplitudes exceeding the switching threshold ($>0.6V$) are applied to the memristor, its internal state variable progressively increases until reaching saturation ($x_{Max} = 1$).  In the absence of write pulses, it gradually decays toward its minimum value ($x_{Min} = 0.1$) due to the intrinsic volatile relaxation of the device. 
Figure \ref{memristor_internal_state}(b) shows the device response to analog input pulse trains, where the normalized input values ($0-0.5$) are linearly scaled to the memristor’s write voltage range ($0.8-1.8V$). Higher-amplitude pulses produce stronger updates in $x$, while lower amplitudes induce smaller changes. 

\subsection{RC with Memristors}
\label{subsec:rc_with_memristors}

%Several memristive devices have been used to perform various RC methods. For example, memristive echo state networks (ESN)~\cite{esn_with_memristor_double_crossbar} have been realized with double crossbar arrays implementing both reservoir and readout. In \cite{memristive_LSM}, analog synaptic circuits have been designed using memristors to realize the recurrent reservoir connections of a liquid state machine (LSM). In \cite{rc_single_memristor_hog}, a single dynamic memristor has been used as the nonlinear reservoir element, enabling mask-free reservoir computing for image recognition. In \cite{rc_with_self_organizing_nanowire_memristors}, the reservoir has been implemented using self-assembled memristive nanowire networks, where the intrinsic conductance dynamics of the memristors provided the recurrent processing for in-materia computation. In this paper, we follow the method of \cite{rc_with_dynamic_memristors}, using multiple discrete volatile memristors as reservoir nodes that provide nonlinear and fading-memory dynamics without arranging them in a crossbar structure.

\question{Para 1: How is RC with memristors done in literature? What method do we use?}

%Due to the variable resistance properties of memristors, they are particularly well-suited for hardware implementation of RC. Crossbars of nonvolatile memristors with random initializations can be used as the reservoir. The input data is projected into the randomly initialized memristors. The summed currents from this crossbar are converted to voltages and fed to two other crossbars. \revision{One of the crossbars represents the recurrent connections, and the other applies an activation function on the summed results from both the first and second crossbars~\cite{esn_with_memristor_double_crossbar, echo_state_gnn}.} This is an example of the hardware implementation of ESN because of the random and recurrent connections in the reservoir (as described in Section \ref{subsec:reservoir_computing}). \revision{Similarly, LSM can be implemented using CMOS spiking neurons for the activations and crossbars of nonvolatile memristors for the synapses that constitute the reservoir~\cite{memristive_LSM, lsm_with_rram_based_analog_digital_accelerator}.} Data is encoded as probabilistic spike trains of voltages. 
%Although these methods employ nonvolatile memristors - whose states need not be read out immediately and which are relatively mature and widely available — their implementation necessitates large crossbar arrays to accommodate high-dimensional reservoirs. Consequently, additional control circuitry is required, leading to a substantial overall area overhead.

\revision{Due to the variable-resistance properties of memristors, several RC families can be mapped to memristive hardware. 
In ESN style realizations, random recurrent weight matrices (as described in Section \ref{subsec:reservoir_computing}) can be implemented using nonvolatile memristor crossbars, enabling efficient in-memory vector–matrix multiplication~\cite{esn_with_memristor_double_crossbar, echo_state_gnn}. 
Additional circuitry is typically required for signal conversion and interfacing (e.g., DAC/ADC and sensing) and for realizing recurrence and nonlinear activation in the overall system. 
Similarly, liquid state machines (LSMs) can be implemented using spiking neurons and memristive synapses, where fixed/random synaptic weights are naturally mapped to resistive-memory arrays~\cite{memristive_LSM, lsm_with_rram_based_analog_digital_accelerator}.
For example, a recent resistive-memory-based LSM hardware–software co-design physically integrates input and recurrent synapses in a $512\times512$ resistive-memory crossbar array and interfaces it with digital circuitry for accumulation and downstream learning~\cite{rram_based_zero_shot_lsm}. 
Such implementations highlight the strengths of LSMs for event-driven, multimodal temporal processing, while also illustrating a common system-level trade-off: realizing high-dimensional recurrent reservoirs often involves substantial synaptic storage and peripheral/interface overhead, whose cost depends on the target reservoir size, precision, and degree of parallelism.}

%Unlike ESN and LSM, DFN requires a single node with memory states. The state of this node at different time steps can be used as the reservoir states instead of the large number of physical nodes in ESN and LSM.
\revision{Unlike ESN and LSM, which construct the reservoir from recurrent networks of physical nodes, DFN generates reservoir states through time-multiplexing of a single nonlinear node with memory.}
Memristors, with their variable resistance, are well-suited for implementing the nonlinear node, and volatile memristors are particularly ideal for DFN because their short-term memory provides the required transient dynamics. The input data at each time step non-linearly modifies the state of the volatile memristor. Since the current state of the memristor depends on the previous states, it exhibits the echo-state property without the need for recurrent connections. The short-term memory of the volatile memristor gives the fading-memory property so that the effect of inputs from the far past diminishes. This method requires a single volatile memristor with several memory states, thus reducing the hardware requirements~\cite{rc_single_memristor_hog}. However, this implementation does not enable parallelism and is time-consuming as the entire input has to be divided and applied in time.

\revision{To improve throughput, instead of a single node, multiple nodes can be used to implement the reservoir, as illustrated in Figure~\ref{all_RC_types_block_diagram}(d). The states of all these nodes at different time steps are used as the reservoir states. This results in a larger reservoir compared to a single-node DFN but requires fewer physical nodes than ESN and LSM. The memristive implementation of this method uses volatile memristors as physical nodes~\cite{rc_with_dynamic_memristors, 2d_reconfig_memristor_for_pdfn}}. Each volatile memristor receives a part of the input or the whole input, depending on the task, in parallel. The intermediate or final states (depending on the task) of each memristor are applied to the readout. This method was first demonstrated using dynamic (volatile) memristors in ~\cite{rc_with_dynamic_memristors}. As mentioned above, to distinguish this method from single-node DFN, we term this method a \textit{parallel delayed feedback network} (PDFN).
%This method enables faster operation compared to regular DFN~\cite{rc_single_memristor_hog} while requiring less hardware compared to ESN\cite{esn_with_memristor_double_crossbar} and LSM~\cite{memristive_LSM}.
Nonvolatile memristors that possess long-term memory are ideal for implementing the readout layer weights that are updated during training and remain fixed during inference~\cite{rc_with_diffusive_memristors}.

\question{Para 2: How exactly does our method work? How do parameters like memristor decay and quantization affect performance?}

To implement memristive PDFN, a set of discrete (non-crossbar) volatile memristors is used as the reservoir. Figure \ref{fig_RC_DFN} illustrates such a design for an image classification task. The input data is converted into temporal pulse trains, and each pulse train is applied to a memristor in the reservoir. The amplitude of the input pulse trains is greater than the write voltage of the volatile memristor. Thus, at each time step, the input pulse modifies the internal state of the volatile memristor to which it is applied. Larger amplitude pulses and pulses in quick succession increase the internal state variable more than lower amplitude pulses and pulses with delays between them, similar to what was shown in Figure \ref{memristor_internal_state}.
As a result, the present value of the internal state variable depends on the history of past inputs (the echo state property), and the volatility weakens the impact of inputs in the far past (fading memory property). 
Hence, for the same initial state variable value, unique pulse trains create unique internal state variable values. 
This happens for every memristor in the reservoir. The internal state variable's value is read by applying read pulses to the memristors, and the resulting current is then used for the readout. The current values are then scaled and applied to a single neural network layer. 

The readout is implemented as either linear or ridge regression, depending on the task. During training, the volatile memristor states are not trained (and cannot be trained); only the readout weights are trained. The trained weights are multiplied by the memristor currents to generate the output.

The method used to convert the input data into pulse trains is crucial to the performance of the reservoir~\cite{preprocessing_methods_for_rc}. The features of the input data must be faithfully represented in the memristor for the readout to correctly classify or predict the output. This is particularly important in cases where the input data is not inherently temporal, as in the case of image classification. Spatial data must be converted into temporal form while preserving its spatial features. This is why preprocessing methods that convert spatial inputs into temporal form while preserving key features are important factors, determining the performance of the system, as we show in section \ref{sec:preprocessing}. 

Another important factor is the properties of the memristive devices. In hardware implementations, analog-to-digital converters (ADCs) need to be used to convert the memristor read currents into digital values for the readout. The resolution of the ADCs constrains the available quantization levels of the memristors. High-resolution ADCs are expensive in terms of area and power and even become the bottleneck of the entire system~\cite{isaac}. Hence, a lower number of memristor quantization levels is desirable. Practical memristors also suffer from device-to-device and cycle-to-cycle variations, and the network performance must be robust to these non-idealities. 

In the following sections, we describe the image recognition task and the preprocessing methods used for it. We then evaluate the performance of memristive PDFN for different preprocessing methods and different memristor parameters. We also analyze the impact of memristor quantization and variability on task performance. 
\revision{Building on the established delay-feedback reservoir framework, we present a systematic co-analysis of preprocessing strategies and volatile-memristor dynamics under practical hardware constraints (quantization and variability), and derive design guidelines for image-classification reservoirs.}

\section{Image Recognition}
\label{sec:image_recognition}

The image recognition task requires categorizing images into one of several categories. In a fully connected deep neural network (DNN), the image is first flattened into a vector of size $m \times n$, where $m$ and $n$ are the rows and columns of the image, respectively. This vector is processed by multiple hidden layers, and the output layer performs the final classification. Although effective, this approach requires a very large number of trainable weights. Additionally, all the weights of the network must be trained using algorithms such as backpropagation. When implemented with nonvolatile memristors, such a network requires many crossbar arrays and substantial peripheral circuitry, leading to high area and power consumption. 

Reservoir computing, and the DFN architecture in particular, avoids this overhead because only the readout layer is trained, and the reservoir itself is fixed. This simple training procedure and small readout layer make the approach especially attractive for temporal processing tasks, offering significant reductions in hardware cost compared to conventional networks. However, image classification is not a temporal task. To use a PDFN for this purpose, the two-dimensional image must be transformed into a temporal signal while preserving the spatial structure needed for recognition. The methods used to convert the input data into a form that the reservoir can process and that enables its key features to be represented within the reservoir are called \textit{preprocessing methods}. 
In our approach, each row of the image is converted into a sequence of voltage pulses that encodes the pixel values (see Figure \ref{fig_RC_DFN}). These pulse sequences are then applied to the volatile memristors in the reservoir as described in Section \ref{subsec:rc_with_memristors}. Directly mapping each row into a long pulse train, however, typically performs poorly because the sequence becomes too long to retain the spatial features effectively. This makes preprocessing essential to ensure that the temporal representation still captures the meaningful structure of the image.

\subsection{Preprocessing Methods}
\label{sec:preprocessing}

For image recognition, several preprocessing methods have been described in literature, including dimensions, sectioning, parity \cite{preprocessing_methods_for_rc}, convolutions \cite{rc_based_convolution}, and histogram of oriented gradient (HOG) \cite{rc_single_memristor_hog}. 
Of these, convolutions and HOG are quite complex and require a large number of parameters for the preprocessing itself, effectively losing the PDFNs area advantage. Sectioning, dimensions, and parity are relatively simple and are explained below.  

\begin{figure}[H]
\centering
\includegraphics[scale=0.4]{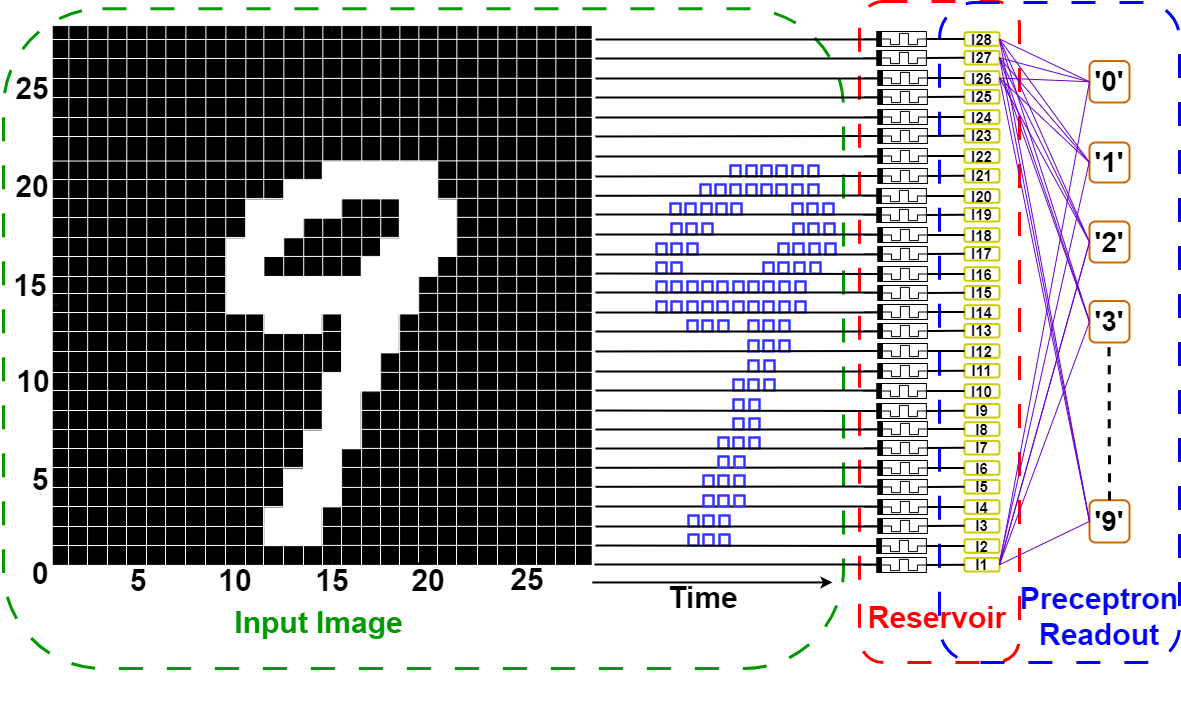}
\caption{Schematic of parallel delayed feedback network (PDFN) reservoir computing (RC). Each row of the binarized input image is converted to a pulse train. The pulse train is then temporally written into the volatile memristors that form the reservoir. After the entire image is written, a read pulse is applied to all volatile memristors, and the obtained currents are rescaled to form the activations of the readout layer. Reproduced from \cite{preprocessing_methods_for_rc}, with permission from IEEE.}
\label{fig_RC_DFN}
\end{figure}

Generally, the image is divided into several parts (potentially overlapping), and the pixels in each part are applied sequentially to a memristor. The simplest division is row-wise, as shown in Figure \ref{fig_RC_DFN}; however, there are several other ways, as we show in this section. The method used to divide the image impacts its representation in the reservoir and subsequently impacts the recognition accuracy, and must therefore be properly chosen. The preprocessing method also determines the number of memristors in the reservoir as listed in Table \ref{table:rc_size} and explained below.

%As mentioned earlier, the input encoding process generates a dynamic state representation within the reservoir. Image recognition requires several memristors—e.g., one for each image row for row-wise division. However, directly feeding raw inputs into the reservoir typically does not yield a linearly separable representation. Therefore, preprocessing techniques are employed to enhance the separability and expressiveness of the reservoir states.

\subsubsection{Dimension}

The concept of \textit{dimension} defines how the two-dimensional spatial information of an input image is converted into temporal sequences suitable for processing by the memristor-based reservoir. In PDFN, the dimensionality determines whether the reservoir receives a one-dimensional (1D) or two-dimensional (2D) temporal representation of the input image. 
%This conversion plays a central role in shaping the diversity and richness of the internal states generated within the reservoir.

\begin{figure}[H]
\centerline{\includegraphics[scale=0.32]{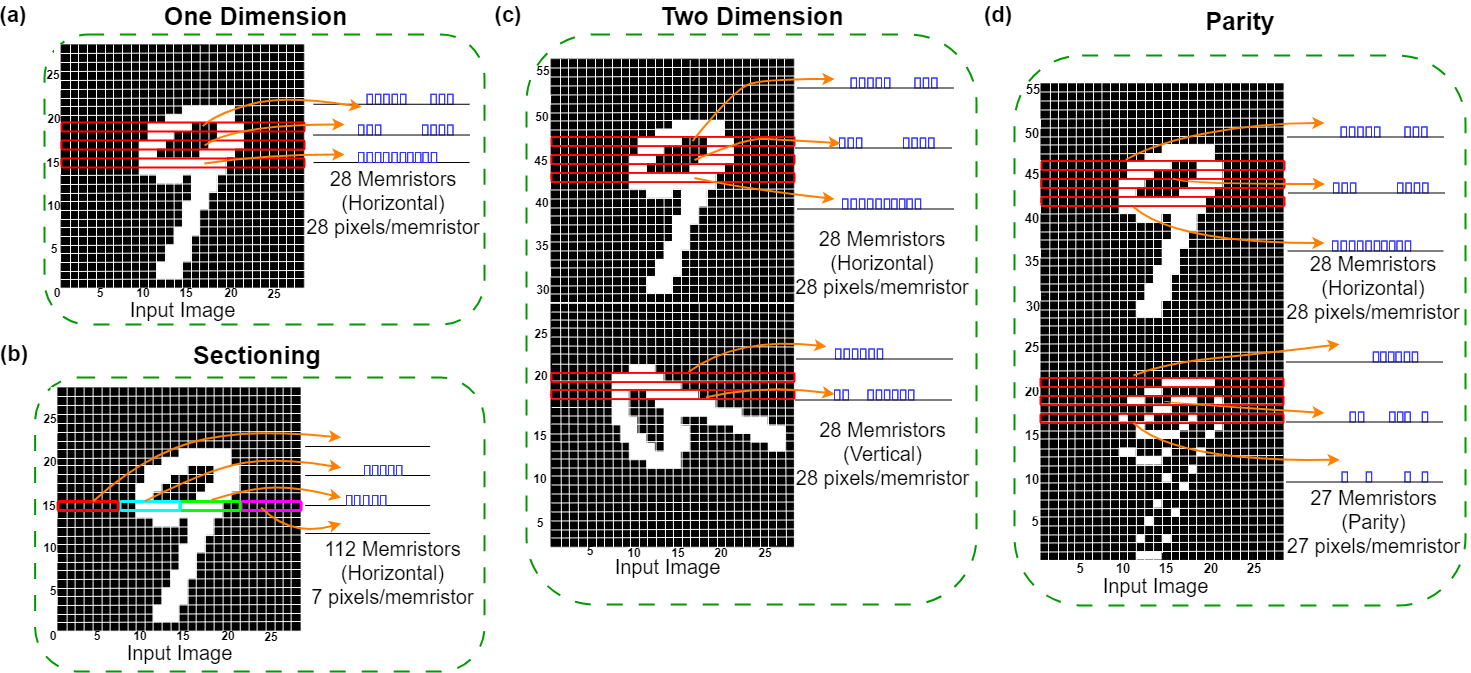}}
\caption{Schematics of preprocessing methods for image processing. (a) One-dimensional preprocessing: each horizontal row of pixels is converted into a pulse train of write voltages and is sequentially injected into a volatile memristor. (b) Input sectioning: only a portion of each row is given to each memristor. In the image shown, four sections are used, and seven pixels are applied to one memristor, improving accuracy by distributing the image representation across multiple devices. (c) Two-dimensional preprocessing: each horizontal row and vertical column of the image is converted into pulse trains of write voltages and sequentially applied to the reservoir's volatile memristors. (d) Parity preprocessing with 1D preprocessing: as in 1D, each horizontal row is converted into a pulse train of write voltages. In addition, the parity operation --- an XOR between adjacent rows --- produces an image with the outline of the digit. This image is also converted into pulse trains of write voltages and applied to separate memristors. Reproduced from \cite{preprocessing_methods_for_rc}, with permission from IEEE.}
\label{fig_RC_preprocessing}
\end{figure}

In the 1D configuration, only the horizontal pixel rows of the image are used to construct the input sequences. Each row is serialized into a temporal stream of voltage pulses, where each pulse corresponds to a pixel value. 
Each row sequence is applied to a distinct volatile memristor, such that an image of size $n \times m$ requires $n$ memristors to represent all horizontal scan lines. As shown in Figure \ref{fig_RC_preprocessing}(a), this approach produces a compact reservoir structure that captures spatial variations along the horizontal axis. The intrinsic temporal decay of the memristor’s internal state enables each device to encode short-term dependencies between successive pixels within a row, effectively translating the spatial structure into temporal dynamics. However, since the vertical correlations between rows are not directly represented, the 1D configuration provides a limited view of the full spatial composition of the image.

The 2D configuration extends this encoding scheme by incorporating both rows and columns of the image. In addition to the $n$ horizontal sequences, the 
$m$ vertical pixel columns are serialized and applied to another set of memristors, resulting in a total of $n+m$ input pulse trains, as illustrated in Figure \ref{fig_RC_preprocessing}(c). This dual encoding offers two orthogonal perspectives of the same image—one capturing horizontal transitions and the other vertical transitions—thereby enriching the temporal diversity of the reservoir. The inclusion of column-based inputs substantially increases (doubles) the number of memristors and consequently the area, but enhances the separability of reservoir states and improves recognition accuracy. Empirically, the 2D configuration has been observed to outperform its 1D counterpart, particularly in datasets such as MNIST, where joining both vertical and horizontal features better defines the image class.

%Increasing the dimension applies the input row-wise as well as column-wise. Overall, allowing the reservoir to perceive the image from a different orientation improves the accuracy. 
%Dimension increase maps each scalar input into a higher-dimensional vector, by applying a $90 \degree$ rotated image along with the original,  which allows the reservoir to perceive the image from a different orientation. 

\subsubsection{Sectioning}

The sectioning configuration provides a systematic method to refine the temporal encoding of spatial information by dividing each input sequence into shorter, localized subsequences. Instead of applying an entire row or column of the image as a single continuous pulse train, the sectioning process partitions the sequence into $k$ smaller segments, each of length $L=m/k$ for an image of width $m$. These subsequences are then applied independently to separate memristors in the reservoir, as illustrated in Figure \ref{fig_RC_preprocessing}(b). This segmentation effectively increases the number of input sequences while reducing the duration of each individual pulse sequence, thereby allowing the reservoir to capture finer temporal dynamics and local spatial structures within the image.

From a hardware point of view, sectioning increases the number of input memristors by a factor of $k$. In a 1D configuration, this leads to a total of $n\cdot k$ memristors, where $n$ is the number of image rows. Similarly, when combined with the 2D configuration, sectioning results in $(n+m) \cdot k$ input sequences, corresponding to all row and column segments. This scaling relationship reflects a deliberate tradeoff: a higher value of $k$ provides greater temporal granularity at the cost of additional memristors. In practice, moderate sectioning factors (e.g., k=4 to 6 for MNIST) yield significant improvements in classification accuracy by enhancing the diversity of reservoir states, while excessively large values may fragment spatial context and reduce overall performance.

The primary motivation behind sectioning is to mitigate the temporal saturation of volatile memristors. When an entire row or column is applied as a long pulse train, the memristor’s internal state may saturate, where additional pulses do not affect the internal state, leading to loss of information as can be seen in the top, black line in Figure \ref{memristor_internal_state}(a). By shortening the effective sequence length, sectioning ensures that each memristor operates within its dynamic range, allowing its intrinsic decay behavior to encode transient correlations more effectively. This results in richer, less redundant internal states, which improve the linear separability of reservoir outputs.

Furthermore, sectioning provides a mechanism for balancing spatial locality and temporal encoding within the reservoir computing framework. Each segment corresponds to a confined spatial region of the image, enabling the system to focus on localized features such as corners, edges, or junctions, while the collective ensemble of segments reconstructs the global structure. This hierarchical representation -- local in each memristor but global in aggregate -- enhances the reservoir’s capacity to process high-resolution patterns and complex spatial arrangements and allows the sections to better represent the 2D features of the image.

\subsubsection{Parity}

The parity configuration introduces an additional preprocessing step designed to enhance the reservoir’s ability to capture spatial transitions and local contrasts within the input image. Unlike the standard dimensional encoding, which converts the raw pixel values directly into temporal pulse sequences, the parity operation derives a new set of input sequences based on the XOR (exclusive OR) relationship between adjacent pixel rows or columns. These parity-derived sequences emphasize edge-like regions and inter-row variations, effectively highlighting structural boundaries that are often critical for pattern discrimination.

When combining parity in a 1D configuration, the preprocessing stage begins by forming an XOR operation between each pair of consecutive rows of the image. For an image consisting of $n$ rows, this produces additional $n - 1$  “parity rows” to the original $n$ rows, resulting in a total of $2n - 1$ input sequences. Each original and parity row is then serialized into a temporal pulse train and applied to an independent volatile memristor, as illustrated in Figure \ref{fig_RC_preprocessing}(d). The parity sequences represent spatial differences rather than absolute intensity, thus encoding transitions between bright and dark regions. As a result, they introduce a complementary feature space that strengthens the separation between classes in the reservoir’s internal state space.

From a hardware and performance standpoint, the parity scheme represents a deliberate balance between input dimensionality and feature expressiveness. It incurs a moderate increase in reservoir size but provides a disproportionate improvement in the system’s ability to distinguish patterns that differ subtly in shape or edge structure. Furthermore, the XOR operation is computationally lightweight and can be implemented efficiently in hardware prior to the reservoir stage, making the approach attractive for low-power edge computing systems. Overall, parity-based preprocessing enriches the temporal representation of spatial data and enhances the separability of reservoir states, leading to improved overall classification performance.

\subsubsection{Summary of Preprocessing Methods}

To sum up, the choice of preprocessing method directly determines how the image is temporally encoded and how many memristors are required in the reservoir. Sectioning divides each sequence into shorter segments to prevent memristor saturation and improve local feature capture; the dimension setting (1D vs. 2D) specifies whether rows alone or both rows and columns are used; and the parity option introduces additional sequences that highlight transitions between adjacent rows or columns. As summarized in Table~\ref{table:rc_size}, combining these options controls the total number of volatile memristors in the reservoir and thus the available spatial and temporal richness, allowing the preprocessing strategy to be tailored to the desired accuracy–area tradeoff.

The table also shows the impact of the different preprocessing methods on latency and on the number of RC memristor writes. Latency is the number of cycles required to process the longest pulse train. Using more sections shortens the pulse trains, reducing the processing latency, but requires more pulse trains (i.e., more RC memristors). The number of RC memristor writes is an indication of the RC energy consumption. Sectioning does not affect the number of writes since the number of processed pixels is independent of the number of sections. In 2D, each pixel belongs to two pulse trains, thus doubling the number of RC memristor writes. In parity, $n-1$ rows are added, increasing the total number of writes to $n \cdot m+(n-1)\cdot m$ in the 1D case, and to $2 \cdot n \cdot m + (n - 1) \cdot m$ in the 2D case.

As noted above, although more accurate preprocessing methods exist, the cost of their additional complexity outweighs the potential benefits.

\section{Evaluation Methodology}
\label{sec:evaluation_methodology}

\question{Para 1: Dataset description and input transformation to pulse trains}
To validate RC capabilities, we use it for the recognition of images from the MNIST dataset \cite{mnist}. MNIST consists of $60,000$ training and $10,000$ testing, grayscale images of handwritten digits of size $28 \times 28$. The spatial organizations of pixels form the features that constitute the images. However, RC needs the input data to be in temporal form. To preserve the spatial features while converting the data into a temporal form, portions of the image are converted into parallel pulse trains as described in Section~\ref{sec:preprocessing}. The number of memristors in the reservoir depends on the chosen preprocessing method. The grayscale pixels are first binarized to reduce computational complexity and emphasize the structural features of the digits. After binarization, the required preprocessing method is applied. Then, each pixel ('1' or '0') is sequentially applied as a pulse train to the memristors of the reservoir.

\question{MATLAB Implementation}

The model of the dynamic memristor was constructed using formulae (\ref{w_update}), (\ref{R(x)}), and (\ref{w_decay}) with parameters from Table \ref{table:dynamic_memristor_parameters} using MATLAB. In this study, parameters are chosen to model a volatile memristor that operates in the nanosecond regime to demonstrate an RC system with throughput comparable to digital hardware. They are not specific to a particular device and are chosen by scaling the parameters of~\cite{model_of_Wox_memristor} to the nanosecond regime. The parameters of the model can be tuned to fit experimental devices of any timescale with similar dynamics as shown in Supporting Information Figure 1. At each time step, a write voltage pulse is applied to the memristor for a duration of $1ns$. If the pixel is '1', a write voltage of $1.5V$ is applied, and if the pixel is '0', $0V$ is applied for the same duration. As described in Section \ref{subsec:memristors}, the memristor updates on the application of a write pulse and leaks on the application of a pulse below $V_{th}$. After all the pulse trains are applied to the memristors, a read pulse of voltage $0.6V$ is applied to all the memristors, and the resulting current represents the internal state of the memristor. These currents are quantized and rescaled by a hardware analog-to-digital converter (ADC) and then applied to the readout layer. Quantization enables evaluation of accuracy in the presence of memristor variability.

The readout layer is implemented as a logistic regression classifier, applying a sigmoid activation element-wise to the linear combination of reservoir states (quantized and rescaled currents) and a weight matrix of dimension ($N\times 10$), where $N$ is the number of reservoir states and $10$ corresponds to the MNIST digit classes. The readout weight matrix is initialized randomly before training. During each epoch, the model performs forward propagation to compute the predicted output, compares it with one-hot encoded target labels, and updates the weights using stochastic gradient descent (SGD) based on the gradient of the binary cross-entropy loss. Training is performed for 500 epochs with a learning rate of 0.02, while only the readout weights are adjusted.

After training, inference is performed on the testing dataset. The images are again converted to pulse trains and applied to the memristor array, after which a read pulse records the final device states. These currents are rescaled and multiplied by the trained readout weights, and a softmax activation function is applied to obtain class probabilities. The predicted class corresponds to the highest probability, and overall test accuracy is computed as the percentage of correctly classified samples among the 10,000 MNIST test images. Memristor resolution is evaluated by varying current quantization from 1 bit to 7 bits (2 to 128 bins).

To simulate the impact of cycle-to-cycle and device-to-device variations, random variations were added to the initialization of the internal state of each memristor, to the decay rate ($\tau$) of each memristor, and to the update and decay equations of the memristor (equations (\ref{w_update}) and (\ref{w_decay})).

Several tests were carried out to evaluate the impact of the preprocessing methods and the memristor parameters -- decay rate ($\tau$), quantization, and variability -- on accuracy.  

\question{Para 3: Readout, logistic regression, and accuracy metric}

\section{Results}
\label{sec:results}

\question{Para 4: Results, effect of device variability, quantization, and decay}

A comprehensive set of experiments was conducted to evaluate how different system and device parameters affect the performance of PDFN memristive reservoir computing on the MNIST dataset. The parameters varied include the preprocessing method, memristor decay rate ($\tau$), quantization level, number of sections, and device variability. Here, we present a meaningful subset of the results to highlight the impact of different parameters on accuracy.  Each figure presents results obtained by isolating one of these parameters while keeping the others fixed, allowing a clear assessment of its individual impact on classification accuracy and reservoir behavior. In addition, a factorial ANOVA was performed to quantify the relative importance of these parameters and to identify which factors and interactions most strongly influence accuracy and robustness.

\subsection{Discussion}

\begin{figure}[H]
\centerline{\includegraphics[scale=2]{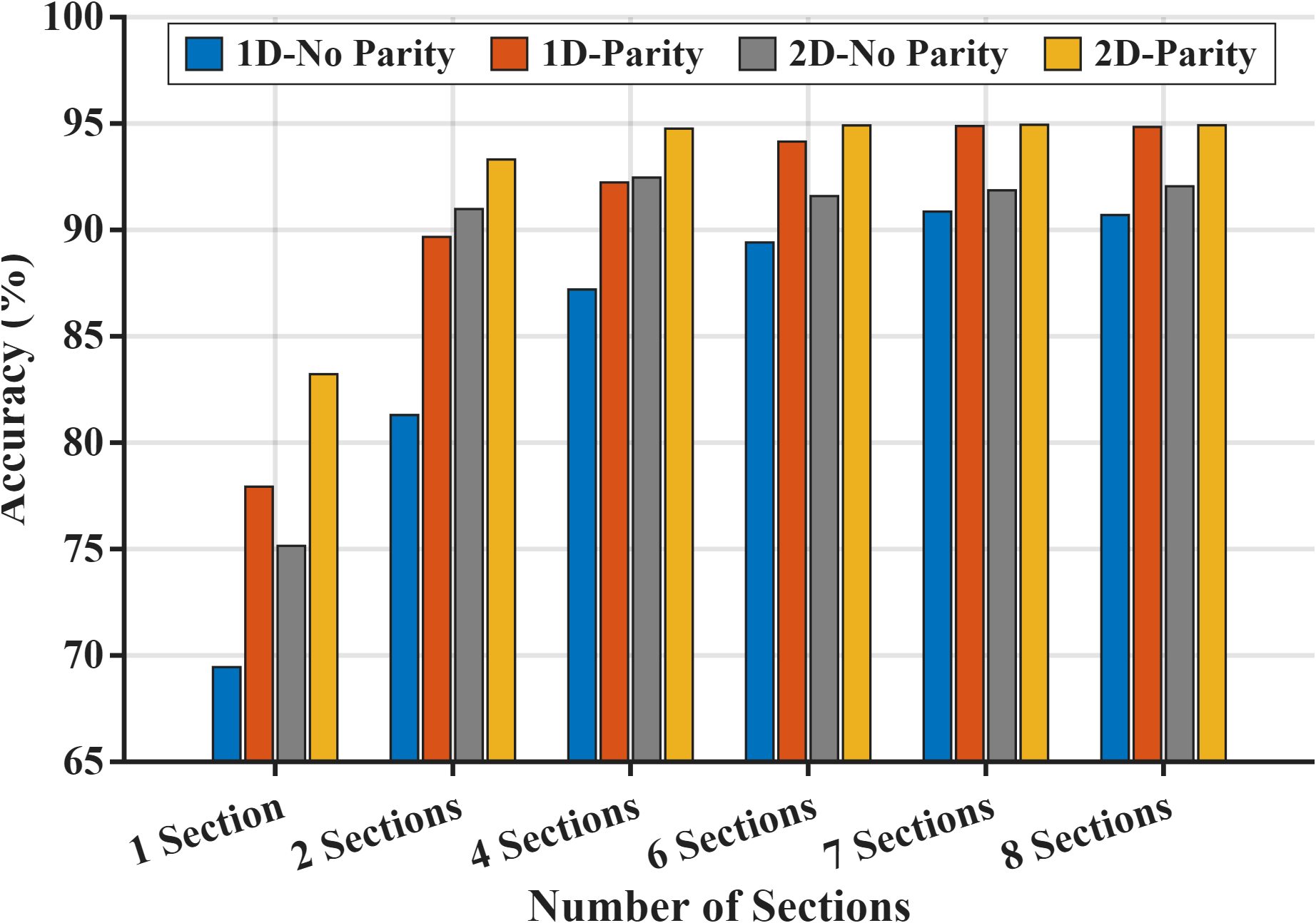}}
\caption{Test accuracy across various preprocessing methods. Data shown for 5-bit quantized memristors with $\tau = 15 ns$. }
\label{preprocessing_results}
\end{figure}

In Figure \ref{preprocessing_results}, different preprocessing methods were implemented, and the corresponding accuracies were measured, for $\tau = 15 ns$ and 5-bit memristor quantization. The preprocessing methods were 1D and 2D, with and without parity for different number of sections. It can be observed that increasing the number of sections per row improves accuracy. This occurs because dividing the image into more sections reduces the number of pixels applied sequentially to each memristor, allowing each device to more accurately capture and represent local image features. Using 2D improves accuracy compared to 1D. Using parity improves accuracy as compared to methods without parity. The amount of improvement in both cases reduces as the number of sections increases. Furthermore, for a larger number of sections ($\geq 6$ sections), parity yields a larger increase in accuracy than an increasing dimension.  

\begin{figure}[H]
\centering 
\includegraphics[scale=2]{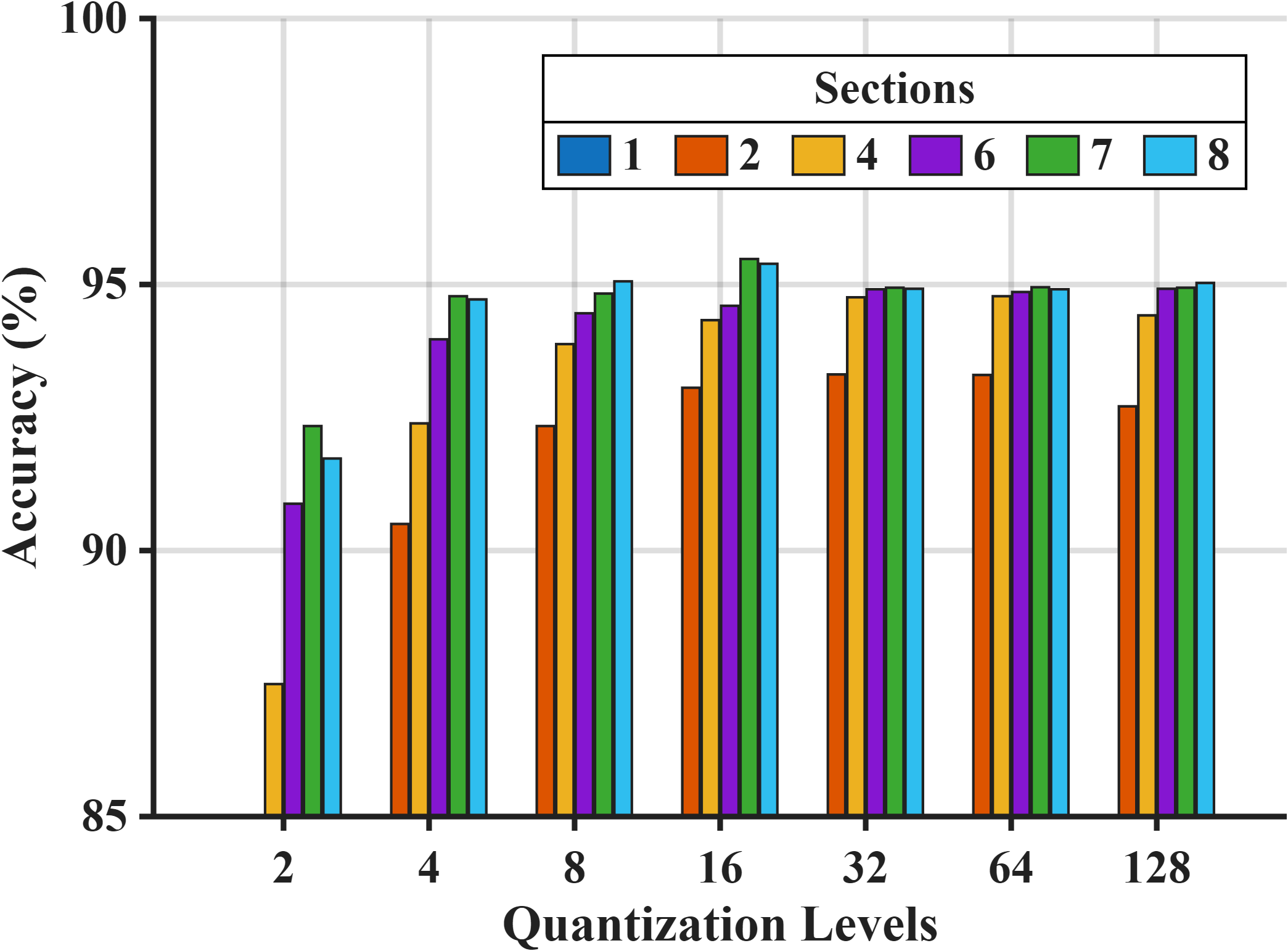}
\caption{Test accuracy versus quantization levels for different section lengths. Data shown is for 2D + parity with $\tau$ = 15 ns.}
\label{Accuracy_vs_Qunatization_all_sections}
\end{figure}

Figure \ref{Accuracy_vs_Qunatization_all_sections} shows the accuracy for different memristor quantization levels and different number of sections, for the preprocessing method 2D $+$ parity and time decay $\tau = 15ns$. Quantization levels were from 1-bit to 7-bit with $1$, $2$, $4$, $6$, $7$ and $8$ sections per row. It can be observed that acceptable accuracy can be achieved even with 2-bit memristor quantization (4 quantization levels). As the number of quantization levels increases, the accuracy increases, but for $\geq32$ levels, more quantization has a diminishing effect. For all quantization levels, $6$ or $7$ sections give optimal accuracy. 

\begin{figure}[H]
\centering 
\includegraphics[scale=2]{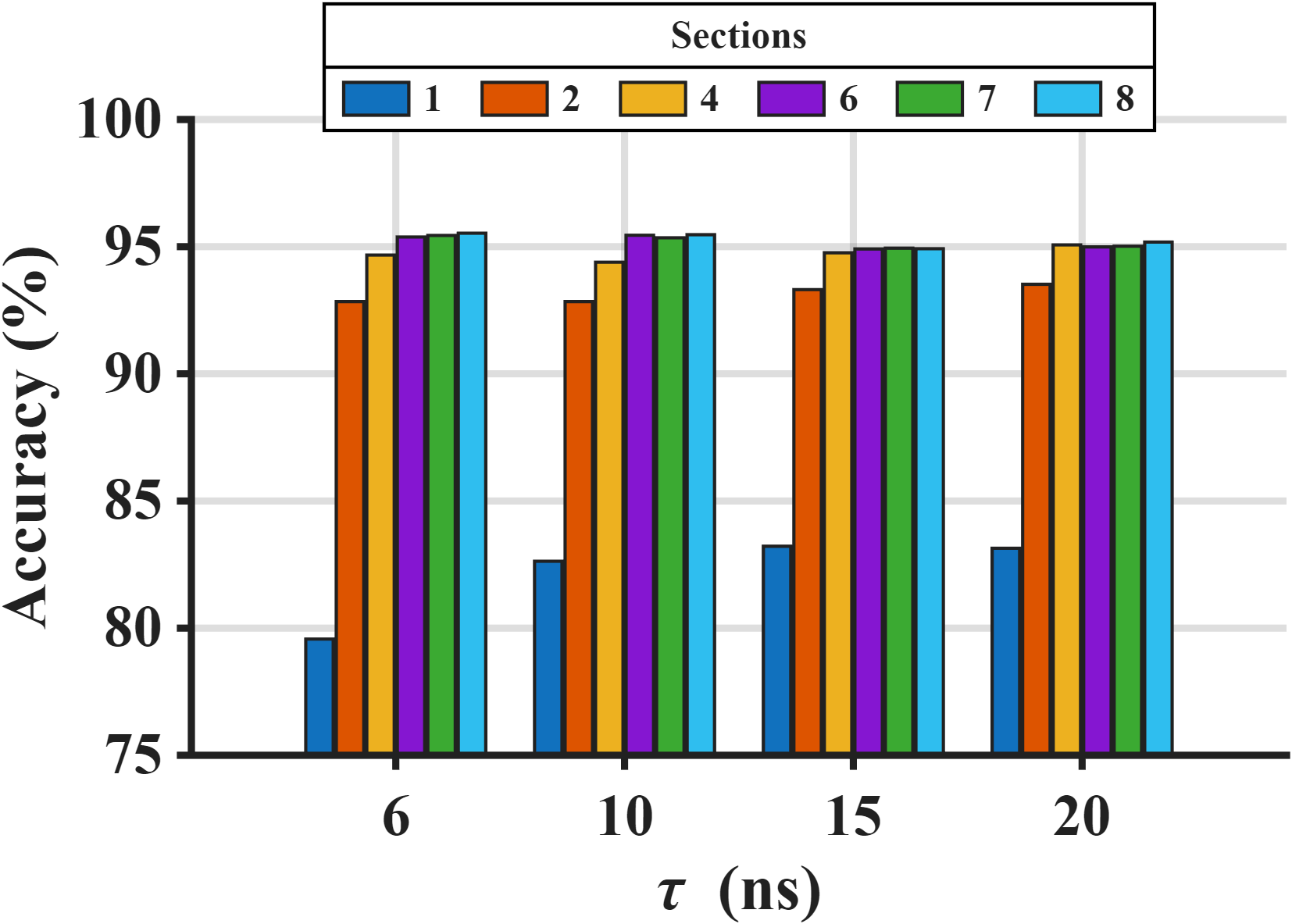}
\caption{Test accuracy versus time decay rate, $\tau$, for different section lengths for 2D + parity with quantization 5-bit.}
\label{Accuracy_vs_tau_all_sections}
\end{figure}

In Figure \ref{Accuracy_vs_tau_all_sections}, different decay time $\tau$ values were tested for different number of sections, for 2D $+$ parity with 5-bit quantization. The $\tau$ values were $6ns$, $10ns$, $15ns$ and $20ns$, with the number of sections as in the previous test. Increasing the decay rate increases accuracy and has a notable impact for fewer sections ($< 4$). For a larger number of sections, the impact of increasing decay rate diminishes. 

\begin{figure}[H]
\centering 
\includegraphics[scale=2]{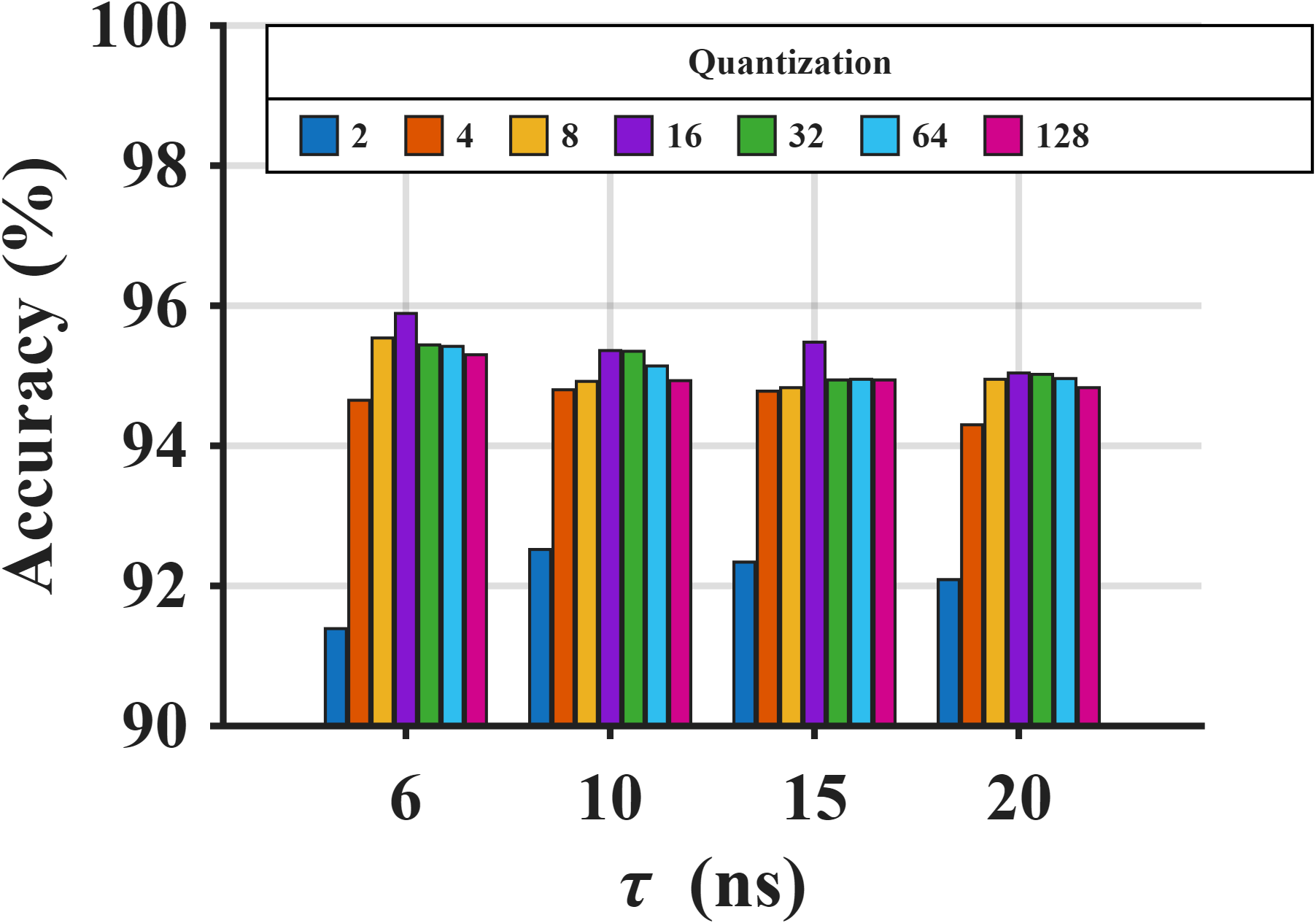}
\caption{Test accuracy versus time decay rate, $\tau$, for different number of memristor quantization levels for 2D + parity with 7 sections.}
\label{Accuracy_vs_tau_all_quantization}
\end{figure}

\begin{figure}[h]
\centering 
\includegraphics[scale=1]{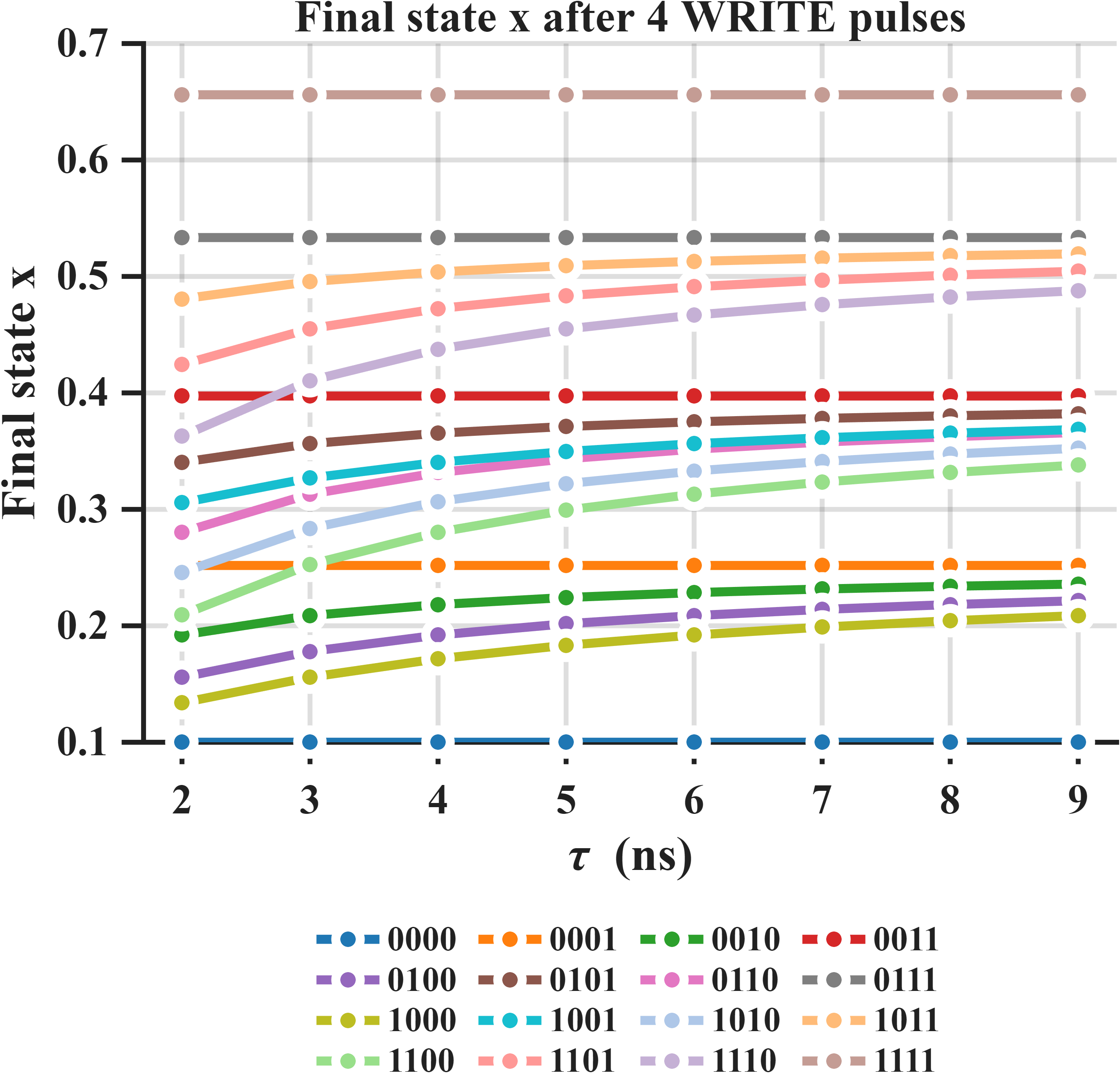}
\caption{Final state $x$ of a single volatile memristor versus decay rate $\tau$ for all possible sequences with four write pulses.}
\label{Final_state_x_vs_tau_for_4_write_pulses}
\end{figure}

In Figure \ref{Accuracy_vs_tau_all_quantization}, different decay rates $\tau$ were tested for different memristor quantization levels, for 2D $+$ parity with $7$ sections. The $\tau$ values were as in Figure~\ref{Accuracy_vs_tau_all_sections}, and the quantization levels were as in Figure~\ref{Accuracy_vs_Qunatization_all_sections}. As an example, it can be seen that $\tau = 6ns$ yields the highest accuracy overall. This is because the input is divided into 7 sections, which means there are 4 pixels per memristor. As a result, $\tau = 6ns$ allows good separation of input pixels. 
This is visualized in the graphs of the final state of a volatile memristor plotted against the decay rate ($\tau$) for all possible input sequences of a given length. Figure~\ref{Final_state_x_vs_tau_for_4_write_pulses} shows the final state, $x$, of a single volatile memristor with different decay rates, $\tau$, for an input sequence with four pulses. At lower $\tau$ values, the memristor is not able to clearly distinguish between different input sequences, as the internal state decays rapidly and becomes dominated by the most recent input pulses; consequently, earlier pixels in the sequence have minimal impact on the final state, leading to significant overlap between sequence representations. If the $\tau$ is too large, the sequences with the same number of '1's converge to the same state, and the memristor acts like a counter, effectively losing the temporal order of the sequence. At intermediate decay rates ($\tau \approx 4$–$6,\mathrm{ns}$ in this case), the final states corresponding to different input sequences are well separated and distributed across multiple quantization levels. Among these, $\tau = 6\mathrm{ns}$ provides the best overall separability, as discussed below.

Because neighboring pixels in natural images are spatially correlated, not all input sequences occur with equal likelihood. As a result, strict separation of all possible sequences is not required; instead, it is sufficient that sequences corresponding to perceptually or statistically distinct image patterns map to different quantization bins. For example, small variations in pixel values—such as slight spatial shifts—that lead to different input sequences but are mapped to the same quantization level can be beneficial, as they promote robustness to minor image perturbations. The $\tau$ value must be chosen such that the pixels of the image applied to the memristor are represented without losing essential information. 
Further discussion about optimal $\tau$ for a given sequence length and quantization is provided in the Supporting Information Note. 

\begin{figure}[H]
\centering 
\includegraphics[scale=1]{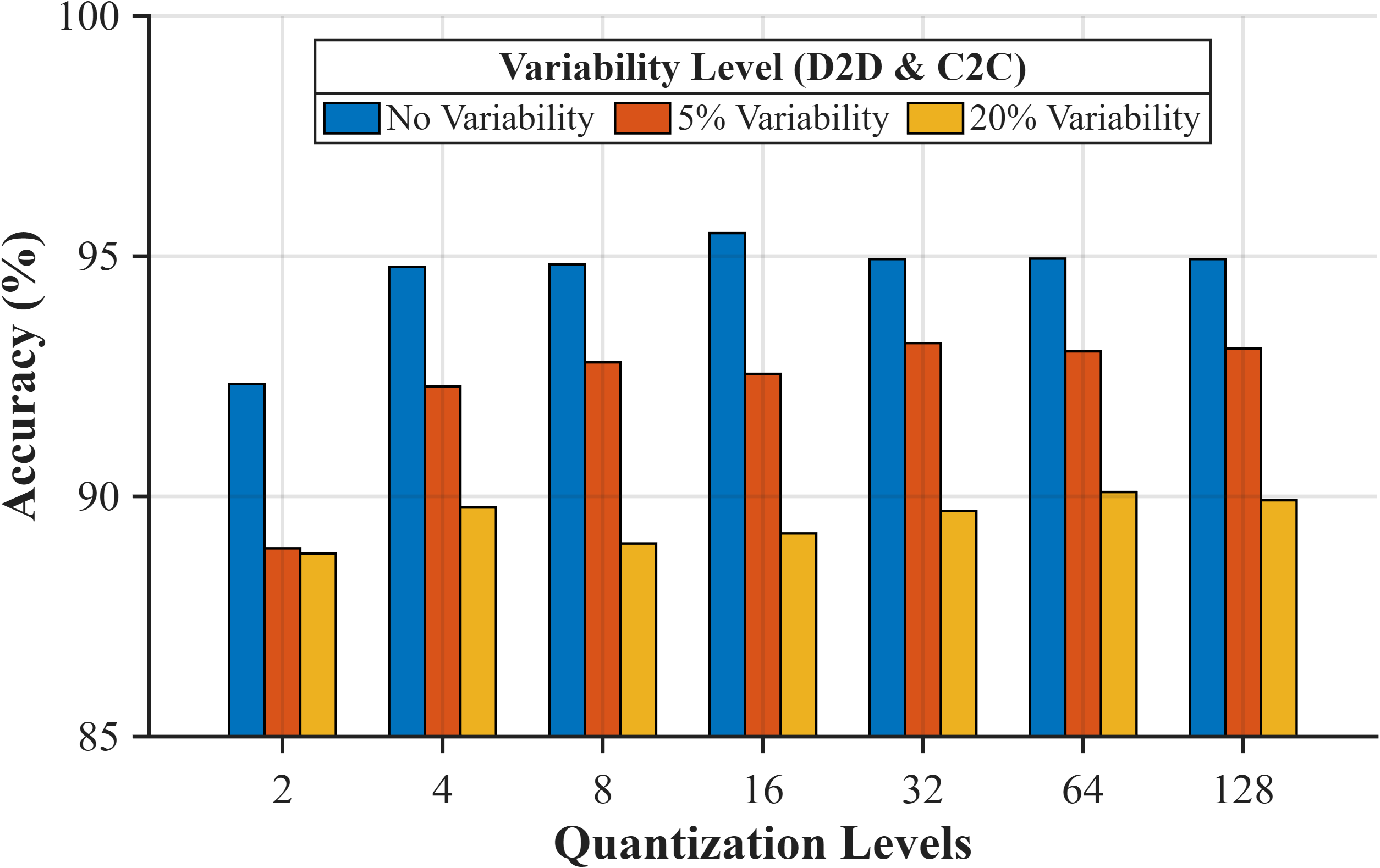}
\caption{Test accuracy versus quantization levels for different variability conditions for 2D + parity with 7 sections and $\tau = 15ns$.}
\label{Accuracy_vs_quantization_for_variability_conditions}
\end{figure}

In Figure \ref{Accuracy_vs_quantization_for_variability_conditions}, different quantization levels were tested for different device-to-device and cycle-to-cycle variability conditions, for 2D $+$ parity with $7$ sections and $\tau = 15ns$. The accuracy was measured for no variability, $5\%$ variability, and $20\%$ variability with the same quantization levels as in Figure~\ref{Accuracy_vs_Qunatization_all_sections}.  It can be observed that adding variability causes a drop in accuracy. This decrease is relatively smaller for $5\%$ variability and is more significant for $20\%$ variability. For $20\%$ variability, the accuracy drops by about $5$ percentage points on average and for $5\%$, it drops by about $2.5$ percentage points on average. 
To isolate the impact of device-to-device variability and validate our analysis, we also performed a Monte-Carlo sweep by jointly varying $\lambda$, $\eta$, $\tau$, and the initialization of the internal state variable $x$ by $5\%$ and $20\%$. The mean, maximum, and minimum accuracy for 30 runs are plotted as error bars in Supporting Information Figure 5. We can observe that for some quantization levels, variability can act as regularization in the training and result in slightly higher accuracy as compared to no variability. Also, it can be seen that while in some runs low quantization (1-bit or 2-bit) can lead to $>90\%$ accuracy, such configurations are unreliable and a quantization of at least 3-bit (8 levels) is needed for robust operation.

For MNIST, the highest accuracy achieved was \MaxAccNoV. This accuracy was achieved for 2D + Parity with 7 sections per row, i.e., 4 pixels per memristor, with $\tau = 6 ns$ and 4-bit quantized memristors (16 quantization levels), as shown in Figure \ref{Accuracy_vs_tau_all_quantization}. This configuration requires $581$ memristors in the reservoir. 
We also measured the effect of quantization of memristor currents on accuracy, and we observed that low-resolution (1 or 2 bit) memristors provide acceptable accuracy (\MaxAccNoVOneBit\ and \MaxAccNoVTwoBit\ respectively). These are both for the 2D + Parity and 7 sections configurations, with decay rate ($\tau$) of 10ns. In both cases, the number of reservoir memristors is $581$ memristors. However, these configurations are unreliable, and quantization of at least 3 bits is needed for robust performance. With $5\%$ variability, the maximum accuracy was \MaxAccVFive\ for the configuration 2D + parity and 8 sections with decay rate of $6ns$ and 3-bit memristor quantization (8 levels). For the same configuration with $20\%$ variability, the maximum accuracy was \MaxAccVTwenty. 

To put the hardware size into perspective, consider a standard fully connected neural network for MNIST recognition. The input layer contains $28 \times 28 = 784$ input pixels. Assume there are two hidden layers, each containing $20$ neurons, and the final layer with $10$ neurons corresponding to the $10$ classes. The total number of trainable weights in this network is $784 \times 20 + 20 \times 20 + 20 \times 10 = 16,280$. In hardware, each weight can be implemented with a nonvolatile memristor. Hence, this network would need $16,280$ nonvolatile memristors.
In a 2D + parity with $8$ sections PDFN with volatile memristors as we have described here, the number of volatile memristors required is $[(28 + 28) + 27] \times 8 = 664$ (Table \ref{table:rc_size}). The number of weights in the readout is $664 \times 10 = 6640$. Each of the readout weights can be implemented with nonvolatile memristors. In total, RC with memristors would require $664$ volatile memristors and $6640$ nonvolatile memristors, which is substantially less than a fully connected network ($16,280$ nonvolatile memristors).

To visualize the representation of the images inside the memristors of the reservoir, we plot the t-distributed stochastic neighbor embedding (t-SNE) of the reservoir currents representing the final states of the memristors after the entire image is applied, and compare them with the plot of the raw MNIST dataset images. The plot is shown in Supporting Information Figure 6. It can be seen that for configurations with high accuracy (eg, $95\%$), the t-SNE shows good separation between the reservoir representations of each digit, whereas the configurations with lower accuracy (eg, $84\%$) have poor separation of the classes in their reservoir representation.

\question{Para 5: How to tune parameters for good performance? Qualitative approach}

\subsection{ANOVA analysis}

To understand how different design choices affect recognition accuracy, we performed a factorial ANOVA, a statistical method that evaluates the influence of several factors at once and determines which ones contribute the most to performance (available in the MATLAB Statistics and Machine Learning Toolbox). This allows us to compare all preprocessing and device parameters on equal footing. The key findings from the ANOVA analysis are summarized below; detailed raw results can be found in the paper's supporting information (Supporting Information Figure 7).

The results show that sectioning has by far the strongest impact on accuracy. Dividing each image into more sections produces consistently higher accuracy, with a particularly large improvement when moving from a single section to multiple sections. Quantization is the next most influential factor: using more quantization levels (e.g., $\geq 16$ levels) substantially improves accuracy by providing a richer representation of each image.

Two additional preprocessing choices also produce clear benefits. Using 2D encoding instead of 1D leads to higher accuracy, demonstrating that preserving the two-dimensional spatial structure produces more informative temporal input streams for the reservoir. Enabling parity encoding also improves accuracy relative to no-parity, because it introduces an additional mixed signal that increases the diversity of internal device states during computation.
The memristor’s leakage time constant $\tau$ has a smaller but still meaningful effect: larger $\tau$ values result in better accuracy for a small number of sections, because a slower decay allows the device to retain information for longer pulse trains.

Several factors also reinforce each other. In particular, sectioning and quantization work synergistically, meaning that fine quantization is most effective when the image is divided into many sections.
Overall, the factorial ANOVA shows that input encoding choices are the primary drivers of accuracy, with sectioning, dimension, and parity contributing the largest improvements, while the device-level time constant and memristor quantization level further shape accuracy and enhance the reservoir’s computational richness.

We next examined how these same architectural and device-level choices influence robustness to device variability. 
Using the same factorial ANOVA framework, we evaluated the change in accuracy under $5\%$ and $20\%$ variability to determine which factors make the reservoir more or less sensitive to noise in the memristor dynamics. Detailed raw results can be found in the paper's supporting information(Supporting Information Figures 8 - 11). 
On average, $5\%$ variability has only a minor effect on accuracy (a reduction of 0.3 percentage points), indicating that the system is generally stable in the low-variability regime. However, larger $20\%$ variability produces a broader spread of outcomes (a reduction of 3.5 percentage points on average), with some configurations degrading only slightly while others suffer substantial accuracy loss. 
The ANOVA reveals that sectioning remains the most influential parameter for robustness. 
The number of sections dominates the variance in $20\%$ variability. It is also the strongest contributor to $5\%$ variability. Quantization is the next most important factor, and its impact grows substantially under $20\%$ variability, reflecting that discretization of memristor state currents into a finite number of levels plays a critical role in how variability propagates through the reservoir. Variability causes a greater reduction in accuracy for higher quantization levels as compared to lower ones. 
Importantly, sectioning and quantization interact strongly: the quantization levels that are robust for one sectioning scheme may be fragile for another, meaning that robustness emerges from their combination rather than from either parameter alone.
The input dimension and parity have smaller main effects, but they meaningfully affect robustness through interactions with sectioning and quantization, while the time constant $\tau$ primarily influences variability tolerance through similar second-order effects. 
Overall, these results show that the same architectural decisions that drive accuracy in the ideal case also govern robustness, most notably the choice of sectioning and quantization, and that variability resilience is best achieved through co-design of input encoding and device-level parameters.

\section{Conclusion}
\label{sec:conclusion}

\question{Summary of key takeaways, insights and learnings}

This paper provides a comprehensive analysis of reservoir computing implemented with volatile memristors, examining how device-level characteristics such as decay rate, quantization, and variability influence system performance. Through analysis of the parallel delayed feedback network (PDFN) architecture, we demonstrate that multiple volatile memristors operating in parallel can effectively capture temporal dependencies, achieving up to \MaxAccNoV\ classification accuracy on MNIST under realistic device constraints, provided the correct design choices are made. These results are comparable with the best existing memristor-based RC implementations. Our results indicate that moderate variability of up to $20\%$ cause a degradation in accuracy but can still be acceptable for memristor resolution of 3-bits or more. Overall, this study highlights volatile memristors as promising building blocks for compact and reliable neuromorphic computing systems.

\medskip
\textbf{Supporting Information}
\label{appendix_A}

Supporting Information is available from the Wiley Online Library or from the author.

\medskip
\textbf{Acknowledgments}

This research has been partially funded by the European Union's Horizon 2020 Research And Innovation Programme FET-Open NEU-Chip under grant agreement No. 964877 and funded by the European Union (ERC, Real-Database-PIM, 101157452). Views and opinions expressed are however those of the author(s) only and do not necessarily reflect those of the European Union or the European Research Council Executive Agency. Neither the European Union nor the granting authority can be held responsible for them.

\medskip
\textbf{Data Availability Statement}

The codes used for the simulations in this study are available on GitHub here:
(\url{https://github.com/rishonadaniels/On_the_role_of_preprocessing_and_memristor_dynamics_in_reservoir_computing}).

\begin{table}[H]
\caption{Parameters of the Dynamic Memristor Model.}
\begin{center}
\begin{tabular}{|c|c|}
\hline
\textbf{Parameter} &  \textbf{Value} \\
\hline
$\alpha$ & $10^{-8} A$ \\
\hline
$\beta$ & $0.5 V^{-1}$\\
\hline
$\gamma$ & $10^{-5} A$ \\
\hline
$\delta$ & $4V^{-1}$\\
\hline
$\lambda$ & $10^3 s^{-1}$\\
\hline
$\eta$ & $8V^{-1}$\\
\hline
$x_{Max}$ & $1$ \\
\hline
$x_{Min}$ &$0.1$\\
\hline
$V_{th}$ & $0.6V$ \\
\hline
$t_{pulse}$ & $1 ns$ \\
\hline
$\tau$ & $1ns - 50ns$\\
\hline
\hline
\end{tabular}
\label{table:dynamic_memristor_parameters}
\end{center}
\end{table}

\begin{table}[H]
\caption{Preprocessing Methods and Reservoir Sizes (assume input image with $n$ rows, $m$ columns, and $k$ sections in both rows and columns)}
\begin{center}
\begin{tabular}{|c|c|c|c|c|}
\hline
\textbf{Dimension} &  \textbf{Parity} & \textbf{No. of Volatile Memristors} & \textbf{Latency} & \textbf{No. of Writes} \\
\hline
1D & No & $n\cdot k$ & $\lceil m/k\rceil$ & $n \cdot m$ \\
\hline
1D & Yes & $[n+(n-1)]\cdot k$  & $\lceil m/k \rceil$ &  $n \cdot m+(n-1)\cdot m$ \\
\hline
2D & No & $(n+m)\cdot k$ & $max(\lceil m/k \rceil, \lceil n/k)\rceil$ & $2\cdot n \cdot m$ \\
\hline
2D & Yes  & $[(n+m)+(n-1)] \cdot k$ & $max(\lceil m/k \rceil, \lceil n/k \rceil) $ & $2\cdot n \cdot m + (n-1) \cdot m$ \\
\hline
% 1D & No & $n\cdot k$ & $\frac{m}{k}$ & $n \cdot m$ \\
% \hline
% 1D & Yes & $[n+(n-1)]\cdot k$  & $\frac{m}{k}$ & $[n+(n-1)]\cdot m$ \\
% \hline
% 2D & No & $(n+m)\cdot k$ & $max(\frac{m}{n}, \frac{n}{k})$ & $(n+m)\cdot m$\\
% \hline
% 2D & Yes  & $[(n+m)+(n-1)] \cdot k$ & $max(\frac{m}{n}, \frac{n}{k})$ & $[(n+m)+(n-1)] \cdot m$ \\

\hline
\end{tabular}
\label{table:rc_size}
\end{center}
\end{table}

\newpage
\vfill
\clearpage

\setcounter{section}{0}
\setcounter{subsection}{0}
\setcounter{subsubsection}{0}
\setcounter{figure}{0}
\setcounter{table}{0}
\setcounter{equation}{0}

\normalfont
\normalsize

\clearpage
\pagestyle{plain}
\thispagestyle{plain}

\setcounter{section}{0}
\setcounter{subsection}{0}
\setcounter{subsubsection}{0}
\setcounter{figure}{0}
\setcounter{table}{0}
\setcounter{equation}{0}

\renewcommand{\thesection}{\arabic{section}}
\renewcommand{\thesubsection}{\thesection.\arabic{subsection}}
\renewcommand{\thesubsubsection}{\thesubsection.\arabic{subsubsection}}
\renewcommand{\thefigure}{\arabic{figure}}
\renewcommand{\thetable}{\arabic{table}}
\renewcommand{\theequation}{\arabic{equation}}

\begin{center}
{\fontsize{18}{22}\selectfont
Supporting Information: On the Role of Preprocessing and Memristor\\
Dynamics in Reservoir Computing for Image Classification\par}
\vspace{1em}
{\normalsize Rishona Daniels, Duna Wattad, Ronny Ronen, David Saad, and Shahar Kvatinsky\par}
\vspace{1.5em}
\end{center}

\normalsize

\captionsetup[figure]{name={Supporting Information Figure}}

\section{Supporting Information Figures}

%\color{red}
%\captionsetup{font={color=red}}

\begin{figure}[!h]
\centering 
\includegraphics[scale=1]{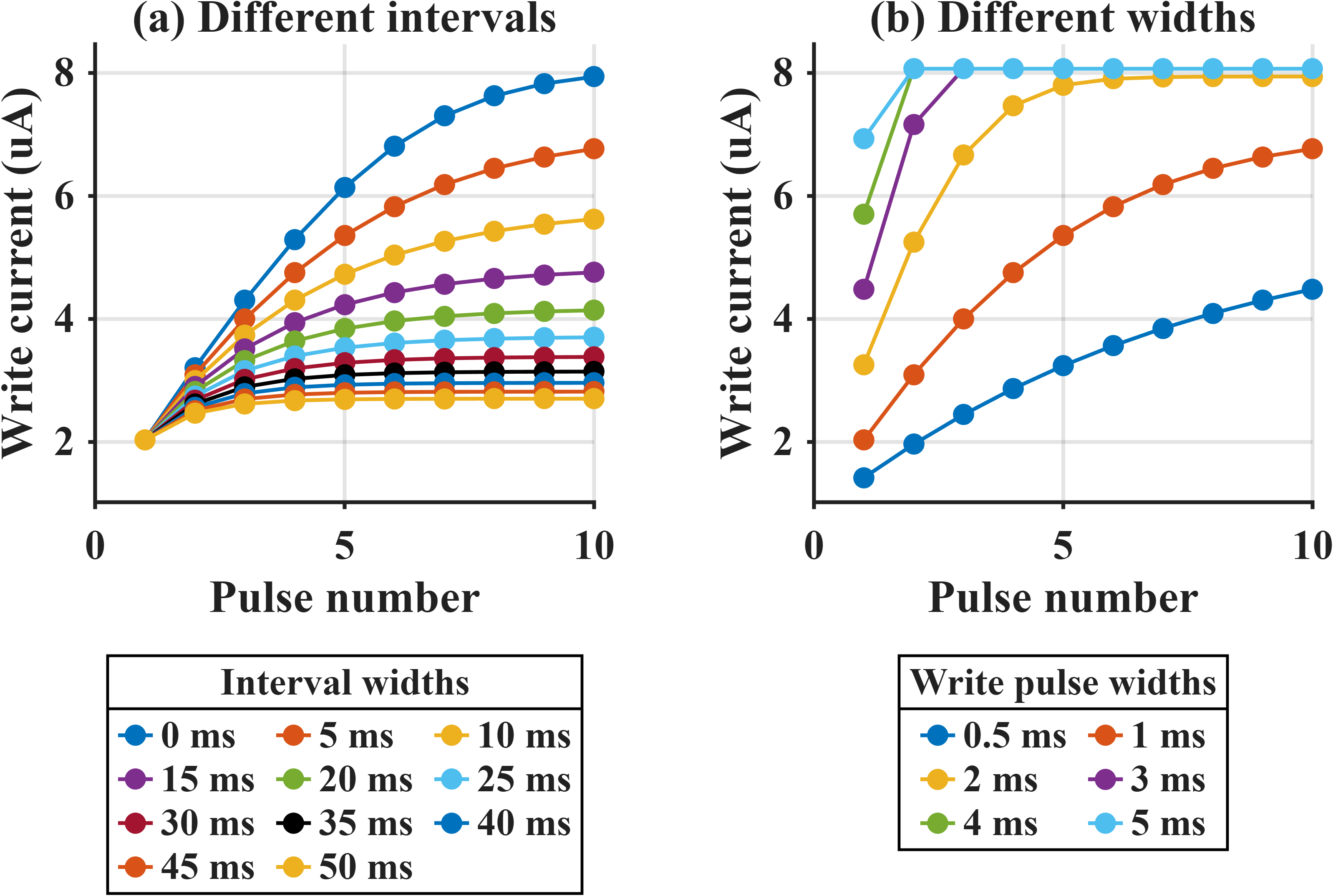}
\caption{Write current versus pulse number of the dynamic memristor model fitted to reproduce the behaviour of the flexible TiO3-WOx3 memristor in \cite{flexible_TiO2_WOx3_memristor}. (a) Current response of the memristor to write voltage pulse trains of fixed width and different intervals between the write pulses. (b) Current response of the memristor to write voltage pulse trains of different widths with fixed interval between the write pulses. }
\label{write_current_vs_pulse_no_for_intervals_and_widths}
\end{figure}

\begin{figure}[!h]
\centering 
\includegraphics[scale=0.5]{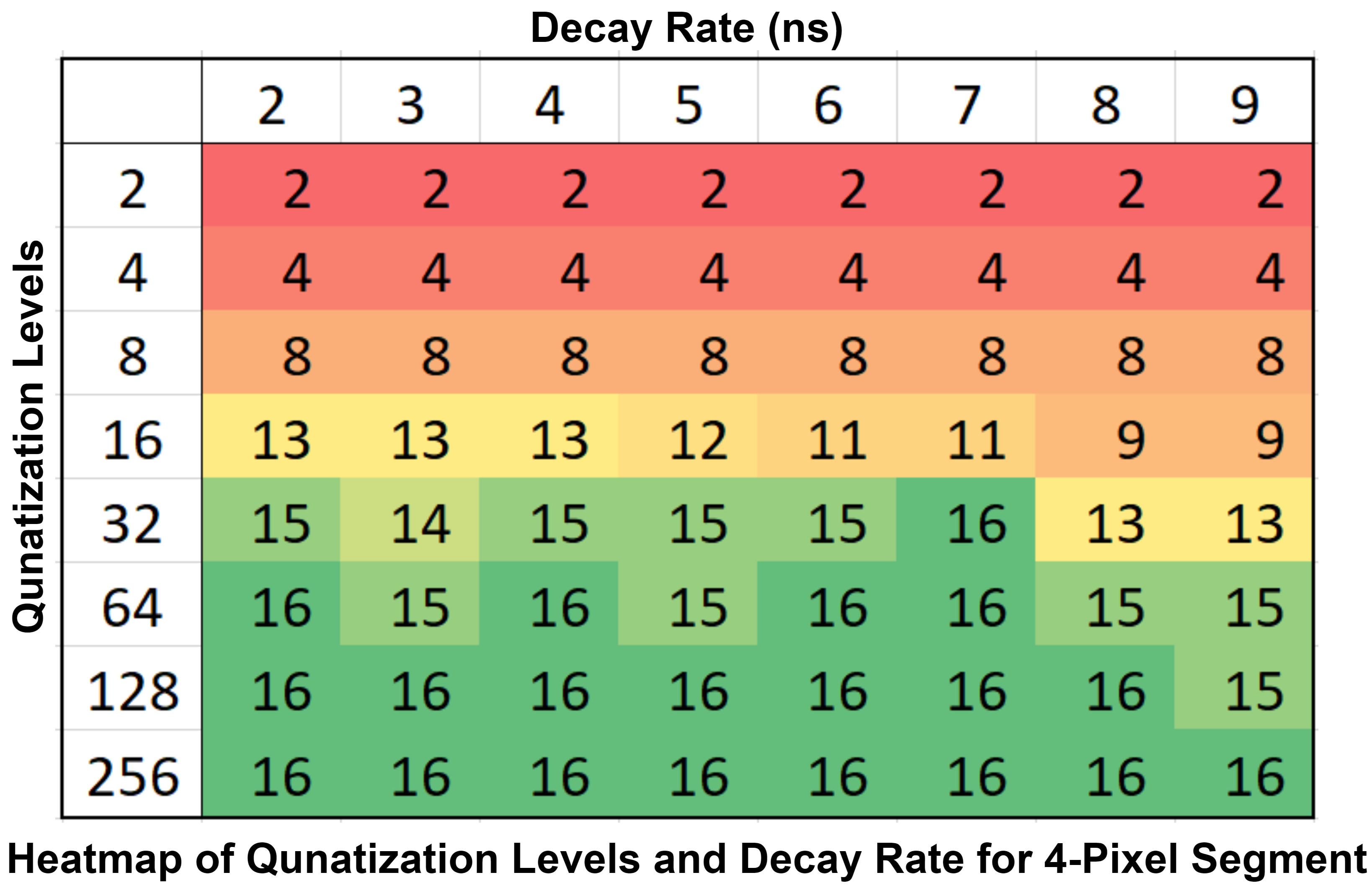}
\caption{Quantization levels versus decay rate for a single volatile memristor for four input write pulses. The numbers in the figure represent the number of memristor states that fall into distinct bins for a given number of quantization levels. To achieve maximum separation for a sequence of four pulses, all 16 possible sequences must be in separate bins. }
\label{optimal_q_and_tau_for_4_pixel_segment}
\end{figure}

    \begin{figure}[!h]
    \centering 
    \includegraphics[scale=1]{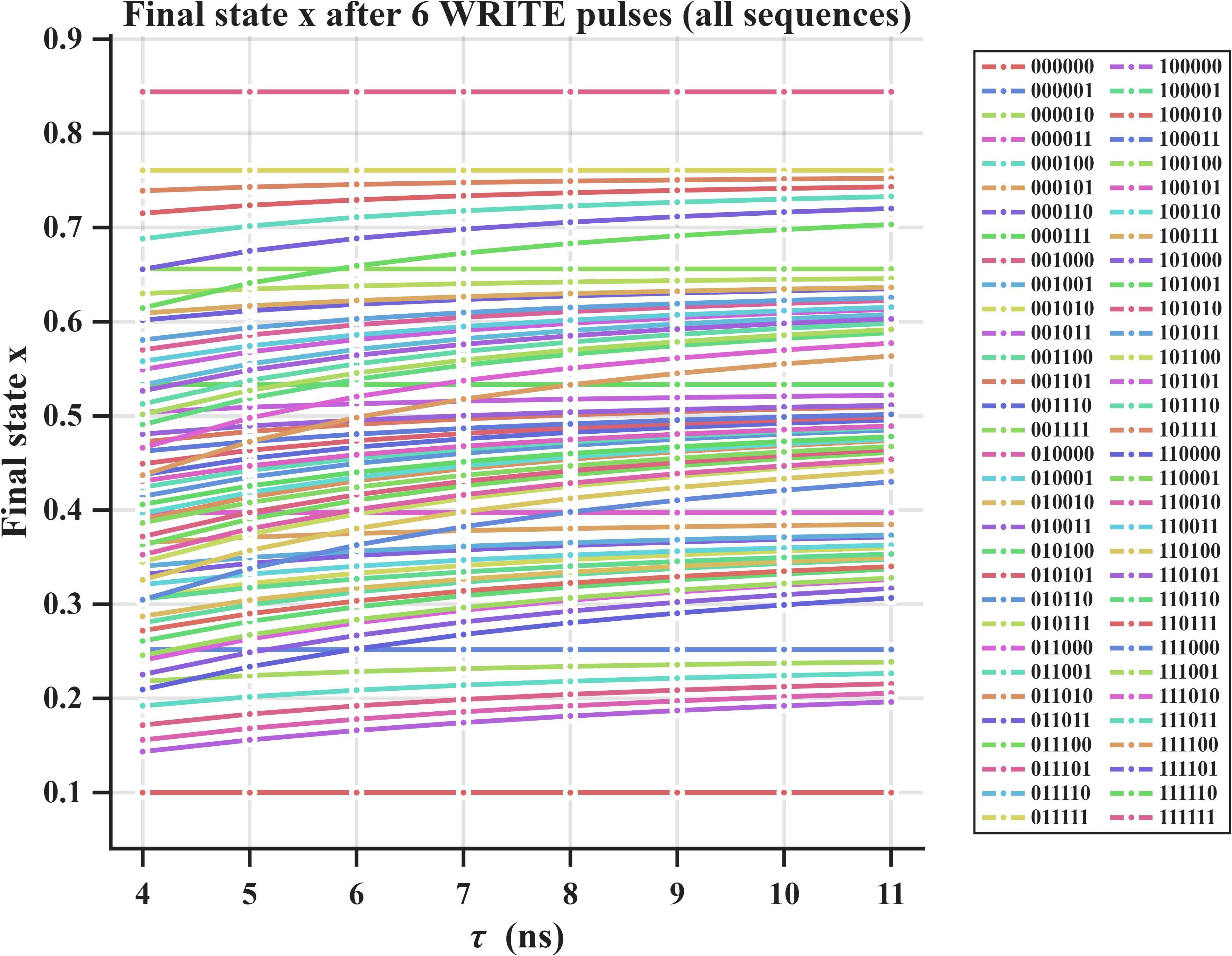}
    \caption{Final state $x$ of a single volatile memristor vs decay rate for all possible sequences with six write pulses.}
    \label{Final_state_x_vs_tau_for_6_write_pulses}
    \end{figure}

    \begin{figure}[!h]
    \centering 
    \includegraphics[scale=0.5]{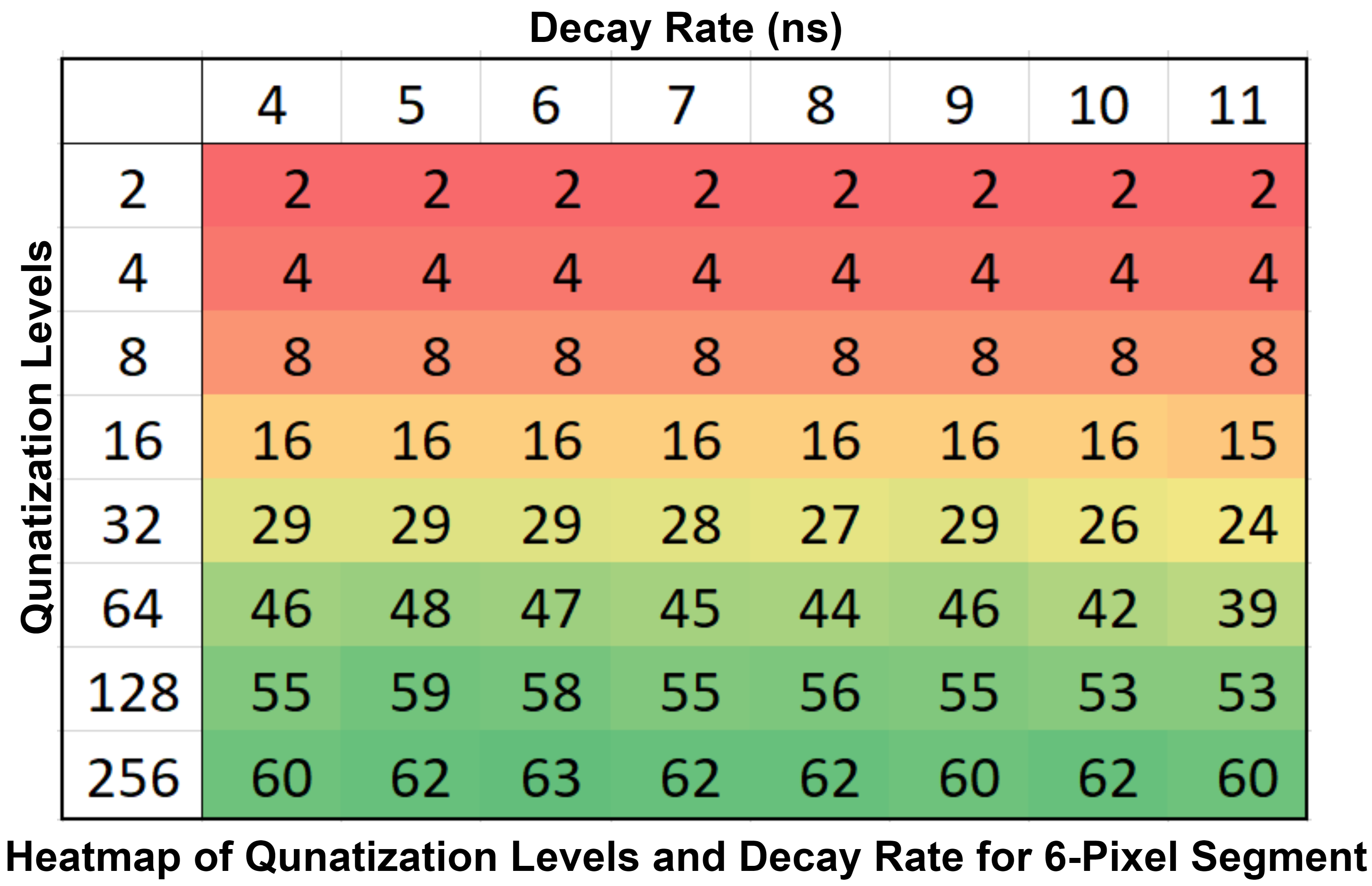}
    \caption{Quantization levels versus decay rate for a single volatile memristor for six input write pulses. The numbers in the figure represent the number of memristor states that fall into distinct bins for a given number of quantization levels.}
    \label{optimal_q_and_tau_for_6_pixel_segment}
    \end{figure}

\begin{figure}[!h]
\centering 
\includegraphics[scale=1]{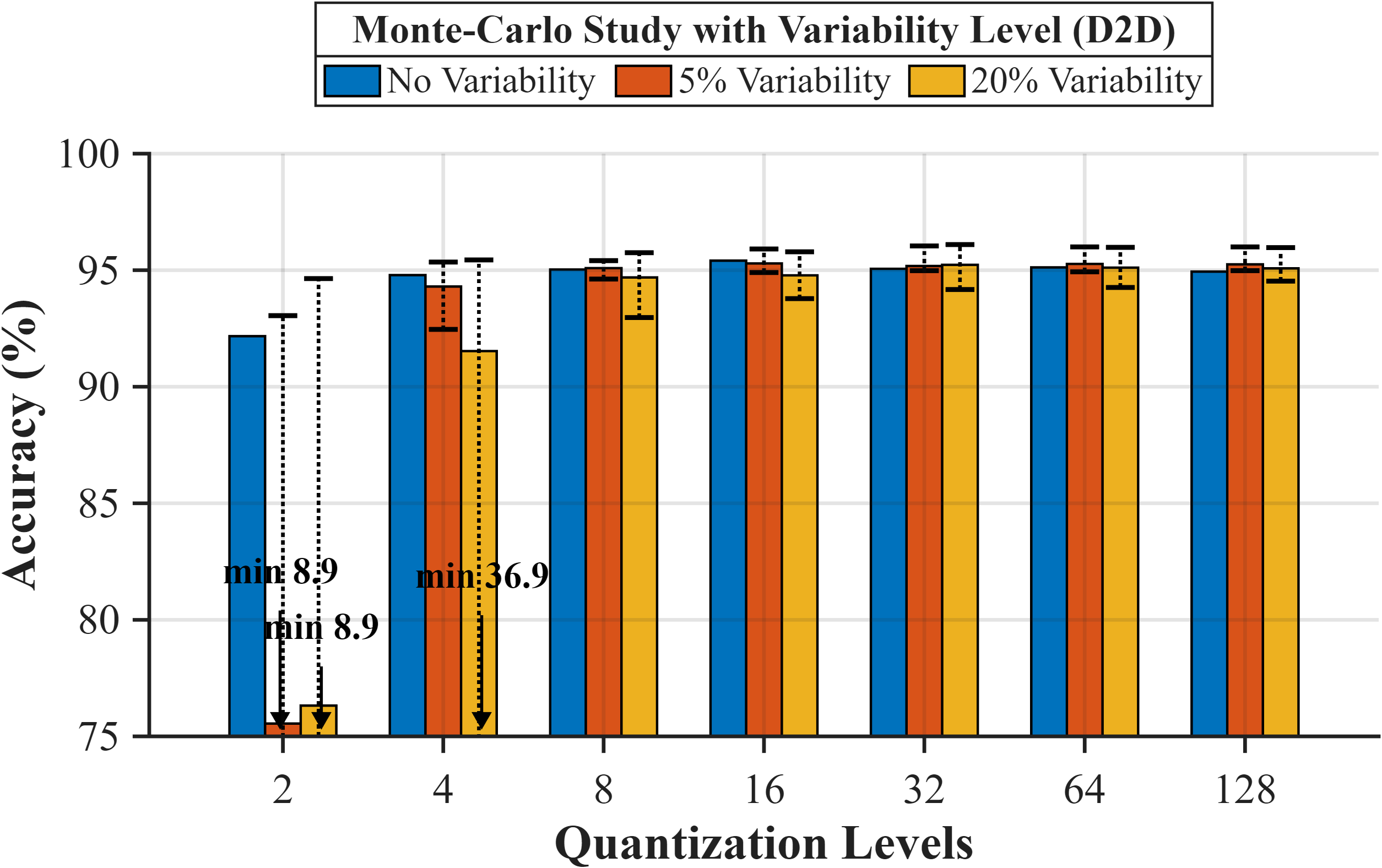}
\caption{Accuracy vs number of quantization levels for 2D + parity with 7 sections and $\tau = 15ns$ for Monte-Carlo sweep of 30 runs for no variability, upto a maximum of $5\%$ variability and upto a maximum of $20\%$ variability with only device-to-device variations.}
\label{accuracy_vs_qunatization_monte_carlo_20_percent}
\end{figure}

\begin{figure}[!h]
\centering 
\includegraphics[scale=1]{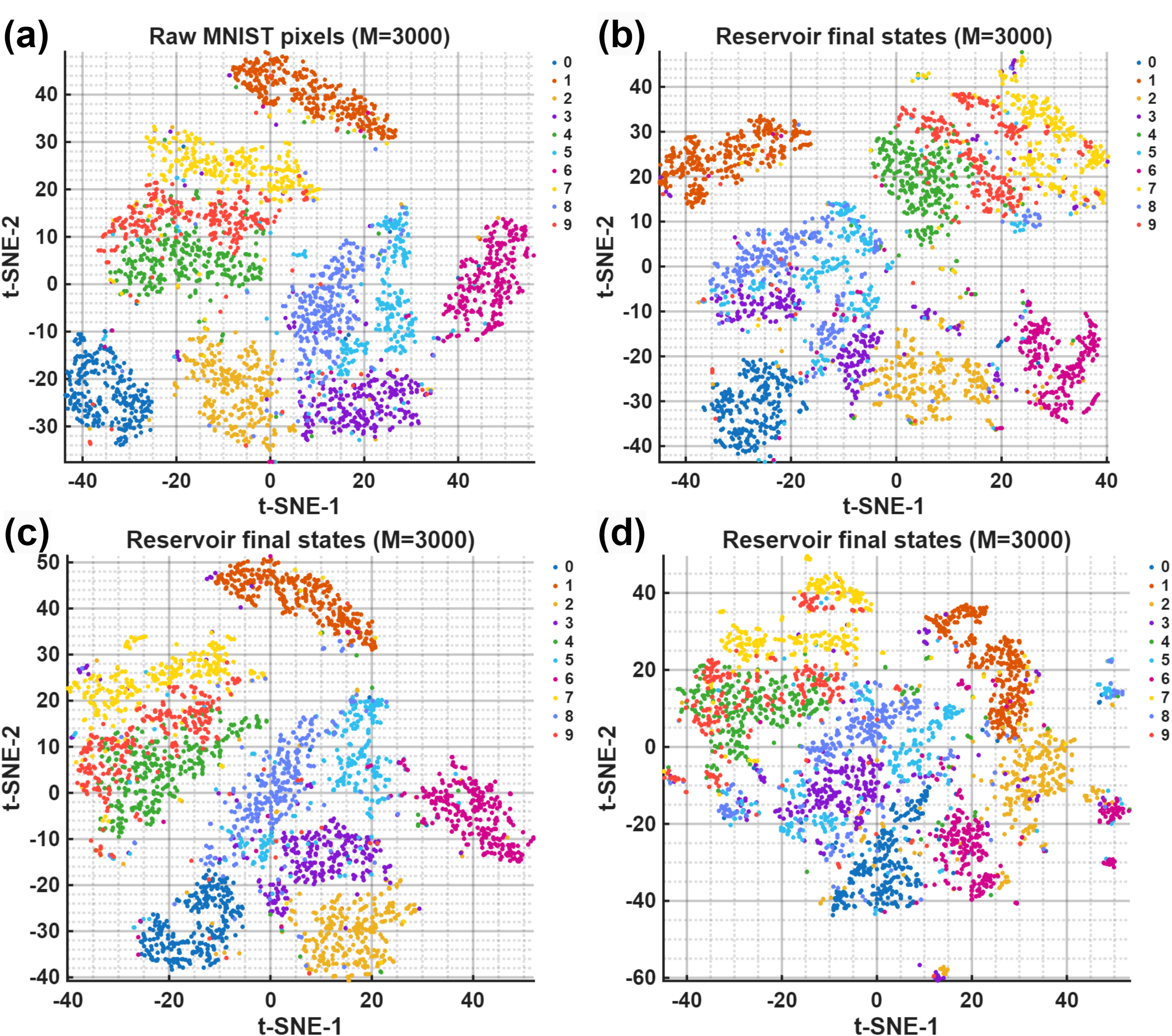}
\caption{t-SNE visualization of (a) raw MNIST dataset, (b) final reservoir currents of configuration 2D, parity, 7 sections, $\tau = 6ns$ and 4-bit quantized currents resulting in an accuracy of $95.88\%$, (c) final reservoir currents of configuration 2D, no parity, 4 sections, $\tau = 10ns$ and 4-bit quantized currents resulting in an accuracy of $90.99\%$, and (d) final reservoir currents of configuration 2D, no parity, 2 sections, $\tau = 10ns$ and 2-bit quantized currents resulting in an accuracy of $84.95\%$.}
\label{t-SNE}
\end{figure}

\captionsetup{font={color=black}}

\begin{figure}[!h]
\centering 
\includegraphics[scale=0.4]{interaction_plots_RC_accuracy.png}
\caption{Pairwise interaction plots showing how accuracy varies across all combinations of preprocessing and device parameters. Each subplot displays the interaction between two factors while averaging over the remaining ones. Strong dependence on sections and quantization is visible through large vertical separations and non-parallel trends, whereas dimension, parity, and $\tau$ exert smaller but still noticeable interaction effects.}
\label{ANOVA_no_variability_interration_plots}
\end{figure}

\begin{figure}[!h]
\centering 
\includegraphics[scale=0.4]{interaction_plots_RC_5_per_v.png}
\caption{Pairwise interaction plots showing how accuracy varies across all combinations of preprocessing and device parameters with $5\%$ variability. Each subplot displays the interaction between two factors while averaging over the remaining ones. Strong dependence on sections and quantization is visible through large vertical separations and non-parallel trends, whereas dimension, parity, and $\tau$ exert smaller but still noticeable interaction effects.}
\label{ANOVA_5_per_variability_interration_plots}
\end{figure}

\begin{figure}[!h]
\centering 
\includegraphics[scale=0.4]{interaction_plots_RC_20_per_v.png}
\caption{Pairwise interaction plots showing how accuracy varies across all combinations of preprocessing and device parameters with $20\%$ variability. Each subplot displays the interaction between two factors while averaging over the remaining ones. Strong dependence on sections and quantization is visible through large vertical separations and non-parallel trends, whereas dimension, parity, and $\tau$ exert smaller but still noticeable interaction effects.}
\label{ANOVA_20_per_variability_interration_plots}
\end{figure}

\begin{figure}[!h]
\centering 
\includegraphics[scale=0.4]{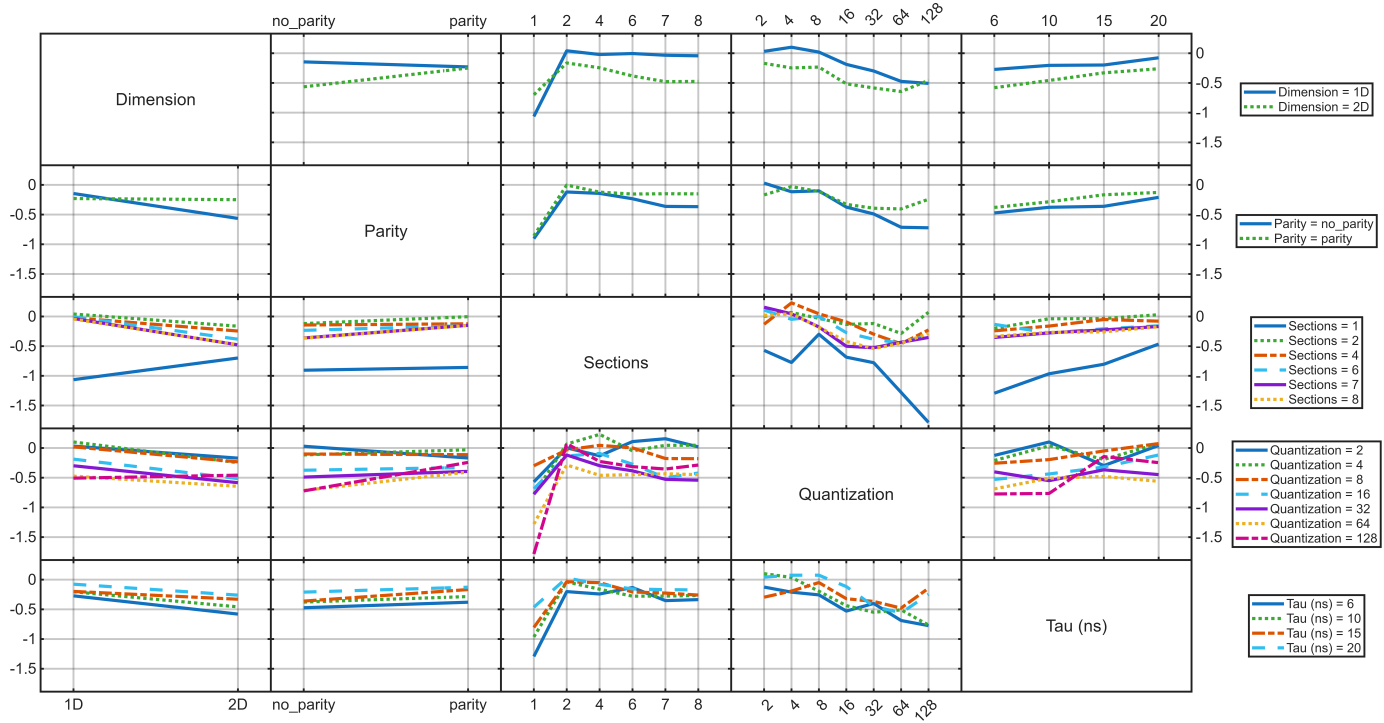}
\caption{Pairwise interaction plots for $\Delta 5$, defined as the percentage change in classification accuracy when introducing $5\%$ device-level variability relative to the baseline (no-variability) accuracy. Positive values indicate a slight improvement under variability, while negative values indicate degradation. In most cases, variability degrades the accuracy as can be seen in the negative values of the delta values. Each subplot shows how $\Delta 5$ depends on one factor across the levels of another factor, with all remaining factors averaged out. Non-parallel curves indicate interaction effects, meaning the influence of one parameter depends on the setting of another. Overall, $\Delta 5$ remains small across all configurations, but interactions—particularly those involving the number of sections and the memristor quantization level—reveal which architectural choices are more or less sensitive to small amounts of variability.}
\label{ANOVA_5_variability_interration_plots}
\end{figure}

\begin{figure}[!h]
\centering 
\includegraphics[scale=0.4]{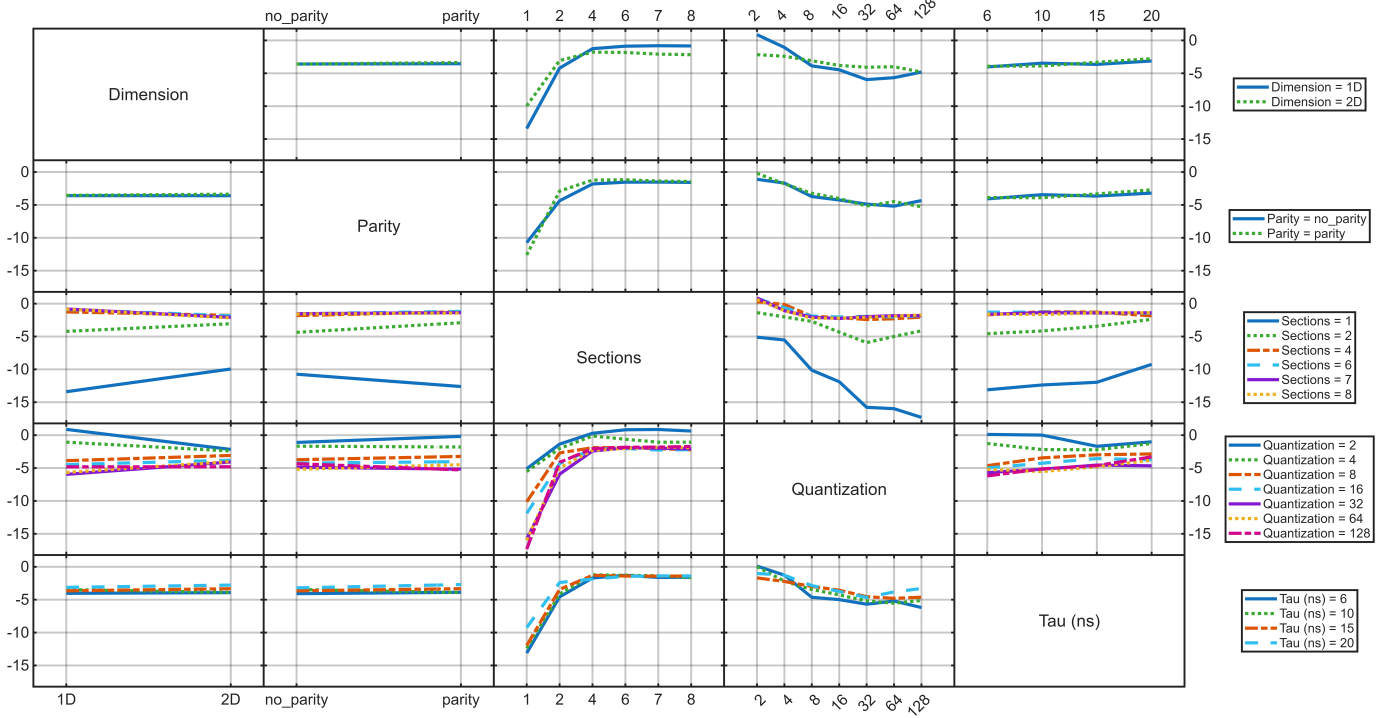}
\caption{Pairwise interaction plots for $\Delta 20$, defined as the percentage change in accuracy when applying $20\%$ device-level variability compared to the baseline (no-variability) accuracy. Larger negative values correspond to greater degradation under variability. In most cases, variability degrades the accuracy, as can be seen in the negative values of the delta values. Each subplot illustrates the interaction between two design parameters (dimension, parity, sections, quantization, and $\tau$), while averaging over the remaining parameters. Strong non-parallel trends show substantial interaction effects, indicating that the impact of variability depends jointly on multiple design choices. Compared to $\Delta 5$, $\Delta 20$ exhibits much larger changes and stronger interactions, especially involving sectioning and quantization, demonstrating that robustness to high variability is highly architecture-dependent.}
\label{ANOVA_20_variability_interration_plots}
\end{figure}

\FloatBarrier

\section{Supporting Information Note}
The optimal quantization level for a given decay rate and a 4-pixel input segment is shown in Supporting Information Figure~\ref{optimal_q_and_tau_for_4_pixel_segment}. When the quantization is too coarse, the internal state of the memristor cannot be resolved with sufficient precision to distinguish between relevant input sequences, leading to a loss of essential information. As the number of quantization levels increases, the final states corresponding to sufficiently distinct and task-relevant sequences are mapped to separate bins, resulting in improved classification accuracy.
For $\tau = 6,\mathrm{ns}$, a quantization of $\geq$ 6 bits (64 levels) is sufficient to uniquely separate all 16 possible input sequences. However, increasing the quantization beyond this point results in a large number of unoccupied or sparsely populated bins. This over-resolution reduces the effective utilization of the available state space and can slightly degrade performance.
Importantly, higher quantization is not always advantageous. Due to the spatial correlation of pixels in natural images, small spatial shifts can map to different input sequences that nonetheless correspond to perceptually similar features. Allowing such closely related sequences to fall into the same quantization bin can be beneficial, as it introduces robustness to small shifts in the input while preserving discrimination between genuinely distinct patterns. Consequently, quantization levels lower than those required for full sequence separation can still yield high accuracy, provided that task-relevant distinctions are maintained.

Similarly, the final state $x$ vs $\tau$ for six-segment pulses and the optimal quantization levels are shown in Supporting Information Figure~\ref{Final_state_x_vs_tau_for_6_write_pulses} and~\ref{optimal_q_and_tau_for_6_pixel_segment} respectively.

%The optimal quantization for a given decay rate for a 4-pixel segment can be seen the table in Supporting Information Figure~\ref{optimal_q_and_tau_for_4_pixel_segment}. For $\tau = 6ns$, $\geq$ 6-bit quantization allows all the possible sequences to be distinguished. However, larger quantization for 16 possible states also means a large number of bins are unoccupied. This results in slightly lower accuracy for larger quantization levels. Also, since the pixels in the image are spatially correlated, certain patterns are more likely than others and as a result, fewer than 16 distinct levels can give high accuracy.


\begin{thebibliography}{1}
\bibliographystyle{IEEEtran}

\bibitem{Jeager_2001_echo_state_approach} H. Jaeger, "The "Echo State" Approach to Analysing and Training Recurrent Neural Networks", \textit{GMD Report 148}, German National Research Center for Information Technology, 2001.

\bibitem{Maass_2003_LSM} W. Maass, T. Natschläger, and H. Markram, "Real-time computing without stable states: A new framework for neural computation based on perturbations," \textit{Neural Computation}, Vol. 14, No. 11, pp. 2531–2560, November 2002.

\bibitem{photonic_rc_for_speech_recognition} E. Picco and S. Massar, "Real-Time Photonic Deep Reservoir Computing for Speech Recognition," \textit{2023 International Joint Conference on Neural Networks (IJCNN)}, Gold Coast, Australia, 2023, pp. 1-7.

\bibitem{hopf_physical_rc_for_sound_recognition} Md R. E. U. Shougat, \textit{et al.}, “Hopf physical reservoir computer for reconfigurable sound recognition,” \textit{Scientific Reports}, Vol. 13, Art. no. 8719, May 2023. 

\bibitem{rc_for_epileptic_seizures} P. Buteneers, \textit{et al.}, “Real-time detection of epileptic seizures in animal models using reservoir computing,” \textit{Epilepsy Research}, Vol. 103, No. 2–3, pp. 124–134, February 2013.

\bibitem{physical_rc_for_robotics} R. Terajima, K. Inoue, K. Nakajima, and Y. Kuniyoshi, “Multifunctional physical reservoir computing in soft tensegrity robots,” \textit{Chaos: An Interdisciplinary Journal of Nonlinear Science}, Vol. 35, No. 8, Art. no. 083111, August 2025.

\bibitem{wind_forecasting_using_rc} Z. Tian, H. Li, and F. Li, “A combination forecasting model of wind speed based on decomposition,” \textit{Energy Reports}, Vol. 7, pp. 1217–1233, November 2021.

\bibitem{stock_market_prediction_using_rc} W.-J. Wang, Y. Tang, J. Xiong, and Y.-C. Zhang, “Stock market index prediction based on reservoir computing models,” \textit{Expert Systems with Applications}, Vol. 178, pp. 115022, September 2021.

\bibitem{img_recog_using_rc} Z. Tong and G. Tanaka, "Reservoir computing with untrained convolutional neural networks for image recognition," \textit{Proceedings of the International Conference on Pattern Recognition (ICPR)}, pp. 1289-1294, Beijing, China, August 2018.

\bibitem{photonic_rc_for_human_action_recognition} E. Picco, P. Antonik, and S. Massar, “High speed human action recognition using a photonic reservoir computer,” \textit{Neural Networks}, Vol. 165, pp. 662–675, August 2023.

\bibitem{ngrc_for_chaotic_systems_prediction} L. Ratas and K. Pyragas, “Application of next-generation reservoir computing for predicting chaotic systems from partial observations,” \textit{Physical Review E}, Vol. 109, p. 064215, June 2024.

\bibitem{experimental_unification_of_rc} D. Verstraeten, B. Schrauwen, M. D’Haene, and D. Stroobandt, “An experimental unification of reservoir computing methods,” \textit{Neural Networks}, Vol. 20, No. 3, pp. 391–403, 2007.

\bibitem{photonic_rc} G. Van der Sande, D. Brunner, and M. C. Soriano, “Advances in photonic reservoir computing,” \textit{Nanophotonics}, Vol. 6, No. 3, pp. 561–576, May 2017.

\bibitem{physical_rc_with_emerging_electronics} X. Liang, \textit{et al.}, “Physical reservoir computing with emerging electronics,” \textit{Nature Electronics}, Vol. 7, pp. 193–206, March 2024.

\bibitem{physical_rc_with_origami} P. Bhovad and S. Li, “Physical reservoir computing with origami and its application to robotic crawling,” \textit{Nature Communications}, 2021.

\bibitem{spintronic_rc} T. Taniguchi, A. Ogihara, Y. Utsumi, and S. Tsunegi, “Spintronic reservoir computing without driving current or magnetic field,” \textit{Scientific Reports}, Vol. 12, no. 10627, 2022.
\bibitem{in_memory_in_sensor_rc_with_memristors_review} N. Lin et al., “In-memory and in-sensor reservoir computing with memristive devices,” \textit{APL Machine Learning}, Vol. 2, No. 1, Art. no. 010901, 2024.

\bibitem{rc_with_dynamic_memristors} C. Du \textit{et al.}, “Reservoir computing using dynamic memristors for temporal information processing,” \textit{Nature Communications}, Vol. 8, No. 1, p. 2204, December 2017.

\bibitem{lukosevicius_jaeger_2009} M. Lukoševičius and H. Jaeger, “Reservoir computing approaches to recurrent neural network training,” \textit{Computer Science Review}, Vol. 3, No. 3, pp. 127–149, 2009.

\bibitem{jaeger_haas_2004} H. Jaeger and H. Haas, “Harnessing nonlinearity: Predicting chaotic systems and saving energy in wireless communication,” \textit{Science}, Vol. 304, No. 5667, pp. 78–80, 2004.

\bibitem{rc_with_self_organizing_nanowire_memristors} G. Milano et al., "In materia reservoir computing with a fully memristive architecture based on self-organizing nanowire networks," \textit{Nature Materials}, Vol. 21, No. 2, pp. 195–202, 2022.
\bibitem{info_processing_in_single_dynamical_system} L. Appeltant, et al., "Information processing using a single dynamical node as complex system," \textit{Nature Communications}, Vol. 2, No. 468, pp. 1–6, 2011.

\bibitem{Chua_1971} L. Chua, "Memristor-The missing circuit element," \textit{IEEE Transactions on Circuit Theory}, Vol. 18, No. 5, pp. 507-519, September 1971. 

\bibitem{Strukov_2008} D. B. Strukov, G. S. Snider, D. R. Stewart, and R. S. Williams, “The missing memristor found,” \textit{Nature}, Vol. 453, No. 7191, pp. 80–83, May 2008.

\bibitem{Chua_Kang_1976} L. O. Chua and S. M. Kang, "Memristive devices and systems," \textit{Proceedings of the IEEE}, Vol. 64, No. 2, pp. 209–223, February 1976.

\bibitem{Chua_2011} L. O. Chua, "Resistance switching memories are memristors,"  \textit{Applied Physics A}, Vol. 102, No. 4, pp. 765–783, March 2011. 

\bibitem{Chua_2014} L. O. Chua, "If it's pinched, it's a memristor," \textit{Semiconductor Science and Technology}, Vol. 29, No. 10, 104001, September 2014.

\bibitem{optically_controlled_MoS2_sec} J. Wang \textit{et al.}, “Optically Controlled MoS2 Phase Conversion Memory-Based In-Sensor Computing Enables Higher Information Security,” \textit{ACS Photonics}, Vol. 12, No. 12, pp. 6946–6956, 2025.

\bibitem{flexible_TiO2_WOx3_memristor}J. Pan \textit{et al.}, “Flexible TiO2-WO3-x hybrid memristor with enhanced linearity and synaptic plasticity for precise weight tuning in neuromorphic computing,” \textit{npj Flexible Electronics}, Vol. 8, Art. No. 70, October 2024.

\bibitem{picosecond_VO2_memristors} S. W. Schmid \textit{et al.}, “Picosecond Femtojoule Resistive Switching in Nanoscale VO2 Memristors,” \textit{ACS Nano}, Vol. 18, No. 33, pp. 21966–21974, 2024.

\bibitem{VTEAM} S. Kvatinsky, M. Ramadan, E. G. Friedman and A. Kolodny, "VTEAM: A general model for voltage-controlled memristors," \textit{IEEE Transactions on Circuits and Systems II: Express Briefs}, Vol. 62, No. 8, pp. 786-790, August 2015.

\bibitem{VTEAM_with_STM} H. Li, D. Kumar, and N. El‑Atab, “A neuromorphic event data interpretation approach with hardware reservoir,” \textit{Frontiers in Neuroscience}, Vol. 18, Art. no. 1467935, November 2024. 

\bibitem{VVTEAM} T. Patni, R. Daniels, and S. Kvatinsky, "V-VTEAM: A compact behavioral model for volatile memristors," \textit{2024 IEEE International Flexible Electronics Technology Conference (IFETC)}, Bologna, Italy, pp. 1-4, 2024.

\bibitem{model_of_Wox_memristor} T. Chang, \textit{et al.}, "Synaptic behaviors and modeling of a metal oxide memristive device," \textit{Applied Physics A: Materials Science \& Processing}, Vol. 102, pp. 857–863, February 2011.

\bibitem{diffusive_memristor} T. Wang \textit{et al.}, "A faithful and compact diffusive memristor model," \textit{IEEE Transactions on Circuits and Systems for Artificial Intelligence}, Vol. 1, No. 2, pp. 141-148, December 2024.

\bibitem{esn_with_memristor_double_crossbar} A. M. Hassan, H. H. Li and Y. Chen, "Hardware implementation of echo state networks using memristor double crossbar arrays," \textit{2017 International Joint Conference on Neural Networks (IJCNN)}, Anchorage, AK, USA, pp. 2171-2177, 2017. 

\bibitem{echo_state_gnn} S. Wang et al., “Echo state graph neural networks with analogue random resistive memory arrays,” \textit{Nat. Mach. Intell.}, Vol. 5, 2023.

\bibitem{memristive_LSM} A. Henderson, C. Yakopcic, S. Harbour and T. M. Taha, "Memristor based circuit design for liquid state machine verified with temporal classification," \textit{2022 International Joint Conference on Neural Networks (IJCNN)}, Padua, Italy, 2022, pp. 1-9. 

\bibitem{lsm_with_rram_based_analog_digital_accelerator} N. Lin \textit{et al.}, “LSMR: Synergy randomness in liquid state machine and RRAM-based analog-digital accelerator,” \textit{Proc. IEEE/ACM Int. Conf. Computer-Aided Design (ICCAD)}, pp. 232:1–232:9, 2024.
\revision{\bibitem{rram_based_zero_shot_lsm} N. Lin \textit{et al.}, "Resistive memory-based zero-shot liquid state machine for multimodal event data learning," \textit{Nature Computational Science}, Vol. 5, pp. 37–47, 2025.}

\bibitem{rc_single_memristor_hog} X. Wu \textit{et al.}, "Nonmasking-based reservoir computing with a single dynamic memristor for image recognition," \textit{Nonlinear Dynamics}, Vol. 112, No. 8, pp. 6663–6678, March 2024.
\revision{\bibitem{2d_reconfig_memristor_for_pdfn} Y. Xia \textit{et al.}, “2D Reconfigurable Memory Device Enabled by Defect Engineering for Multifunctional Neuromorphic Computing,” \textit{Advanced Materials}, Vol. 36, Art. no. 2403785, 2024.}

\bibitem{rc_with_diffusive_memristors} R. Midya \textit{et al.}, “Reservoir Computing Using Diffusive Memristors,” \textit{Advanced Intelligent Systems}, Vol. 1, p. 1900084, 2019.

\bibitem{preprocessing_methods_for_rc} R. Daniels \textit{et al.}, "Preprocessing Methods for Memristive Reservoir Computing for Image Recognition," \textit{Proceedings of the IEEE International Conference on Metrology for eXtended Reality, Artificial Intelligence and Neural Engineering}, pp. 1-6, October 2025.

\bibitem{isaac} A. Shafiee \textit{et al.}, "ISAAC: A Convolutional Neural Network Accelerator with In-Situ Analog Arithmetic in Crossbars," \textit{2016 ACM/IEEE 43rd Annual International Symposium on Computer Architecture (ISCA)}, Seoul, Korea (South), 2016, pp. 14-26.

\bibitem{rc_based_convolution} Y. Tanaka and H. Tamukoh, "Reservoir-based convolution,"  Special Section on Nonlinear Science Workshop on the Journal, \textit{Nonlinear Theory and Its Applications, IEICE}, Vol. 13, No. 2, pp. 397–402, 2022.

%\bibitem{NARMA} A. F. Atiya and A. G. Parlos, "New results on recurrent network training: unifying the algorithms and accelerating convergence," \textit{IEEE Transactions on Neural Networks}, Vol. 11, No. 3, pp. 697-709, May 2000. 

\bibitem{mnist} Y. Lecun, L. Bottou, Y. Bengio and P. Haffner, "Gradient-based learning applied to document recognition," \textit{Proceedings of the IEEE}, Vol. 86, No. 11, pp. 2278-2324, November 1998.

\bibitem{neural_network_zoo} Asimov Institute, “The neural network zoo,” 2019. [Online]. Available: https://www.asimovinstitute.org/neural-network-zoo/. Accessed: November 26, 2025.



\end{thebibliography}

\begin{thebibliography}{00}

\bibitem[S1]{flexible_TiO2_WOx3_memristor}J. Pan \textit{et al.}, “Flexible TiO2-WO3-x hybrid memristor with enhanced linearity and synaptic plasticity for precise weight tuning in neuromorphic computing,” \textit{npj Flexible Electronics}, Vol. 8, Art. No. 70, Oct. 2024.

\end{thebibliography}
\end{document}